\documentclass[a4paper,fleqn]{cas-dc}

\usepackage[authoryear]{natbib}
\usepackage{algorithm}
\usepackage{amsmath}
\usepackage{algpseudocode}
\usepackage{subfig}
\usepackage{siunitx}
\usepackage{bm}
\usepackage{lineno} 
\def\tsc#1{\csdef{#1}{\textsc{\lowercase{#1}}\xspace}}
\tsc{WGM}
\tsc{QE}
\begin{document}
\let\WriteBookmarks\relax
\def\floatpagepagefraction{1}
\def\textpagefraction{.001}

\shorttitle{H. Li et~al./ Expert Systems with Applications}
\shortauthors{H. Li et~al.}

\title [mode = title]{DK-STN: A Domain Knowledge Embedded Spatio-Temporal Network Model for MJO Forecast}  

% \tnotemark[<tnote number>] 


\author[1,2]{Hongliang Li}[style=chinese, orcid=0000-0003-4377-0550]



\ead{lihongliang@jlu.edu.cn}

%\ead[url]{www.cvr.cc, cvr@sayahna.org}

\credit{Conceptualization, Supervision, Writing – review $\&$ editing, Project administration}

%\affiliation[1]{organization={College of Computer Science and Technology},addressline={Jilin University}, city={Changchun},
    % citysep={}, % Uncomment if no comma needed between city and postcode postcode={130012}, 
    % state={},country={China}}

%\address{College of Computer Science and Engineering, University Name, Chongqing 400054, China}
\address[1]{College of Computer Science and Technology, Jilin University, Changchun, 130012, China}

\author[1]{Nong Zhang}[style=chinese, orcid=0000-0001-9753-3556]
\ead{zhangnong21@mails.jlu.edu.cn}

\credit{Conceptualization, Methodology, Software, Writing – original draft, Visualization}

\author[1]{Zhewen Xu}[style=chinese, orcid=0000-0001-8309-6084]
%\fnmark[3]
\ead{zwxu20@mails.jlu.edu.cn}

\credit{Methodology, Software, Writing – review $\&$ editing, Visualization}

%\affiliation[2]{organization={Key Laboratory of Symbolic Computation and Knowledge Engineering of the Ministry of Education},
    % addressline={},     city={Changchun},
    % citysep={}, % Uncomment if no comma needed between city and postcode    postcode={130012},     country={China}}

\address[2]{Key Laboratory of Symbolic Computation and Knowledge Engineering of the Ministry of Education, Changchun, 130012, China}

\author[1]{Xiang Li}[style=chinese]
%\fnmark[]
\ead{lxiang@jlu.edu.cn}
\credit{Investigation}
% Fifth author
\author[3]{Changzheng Liu}[style=chinese, orcid=0000-0002-8789-1840]
\cormark[1]
\ead{czliu@cma.gov.cn}
\credit{Data curation, Investigation}
% Sixth author
\author[3]{Chongbo Zhao}[style=chinese]
\ead{zhaochongbo@cma.gov.cn}
\credit{Data curation, Investigation}
% Seventh author
\author[3]{Jie Wu}[style=chinese]
\ead{wujie@cma.gov.cn}
\credit{Data curation, Investigation}

%\affiliation[3]{organization={China Meteorological Administration Key Laboratory for Climate Prediction Studies},    addressline={National Climate Center},     city={Beijing},
    % citysep={}, % Uncomment if no comma needed between city and postcode    postcode={100081},     country={China}}

\address[3]{China Meteorological Administration Key Laboratory for Climate Prediction Studies, National Climate Center, Beijing, 100081, China}

\cortext[cor1]{Corresponding author}

\begin{abstract}
Understanding and predicting the Madden-Julian Oscillation (MJO) is fundamental for precipitation forecasting and disaster prevention. To date, long-term and accurate MJO prediction has remained a challenge for researchers. Conventional MJO prediction methods using Numerical Weather Prediction (NWP) are resource-intensive, time-consuming, and highly unstable (most NWP methods are sensitive to seasons, with better MJO forecast results in winter). While existing Artificial Neural Network (ANN) methods save resources and speed forecasting, their accuracy never reaches the 28 days predicted by the state-of-the-art NWP method, i.e., the operational forecasts from ECMWF, since neural networks cannot handle climate data effectively. In this paper, we present a Domain Knowledge Embedded Spatio-Temporal Network (DK-STN), a stable neural network model for accurate and efficient MJO forecasting. It combines the benefits of NWP and ANN methods and successfully improves the forecast accuracy of ANN methods while maintaining a high level of efficiency and stability. We begin with a spatial-temporal network (STN) and embed domain knowledge in it using two key methods: (i) applying a domain knowledge enhancement method and (ii) integrating a domain knowledge processing method into network training. We evaluated DK-STN with the 5th generation of ECMWF reanalysis (ERA5) data and compared it with ECMWF. Given 7 days of climate data as input, DK-STN can generate reliable forecasts for the following 28 days in 1-2 seconds, with an error of only 2-3 days in different seasons. DK-STN significantly exceeds ECMWF in that its forecast accuracy is equivalent to ECMWF's, while its efficiency and stability are significantly superior.
\end{abstract}

% \includegraphics{figs/grabs.pdf}
%\item We explored long-term predictions for the Madden-Julian Oscillation (MJO).
%\item We propose a Domain Knowledge embedded Spatio-Temporal Network model (DK-STN).
%\item DK-STN can predict long-term MJOs with few circulation variables.
%\item Our results outperform existing methods and have higher efficiency and stability.

% Keywords
%climate prediction, domain knowledge, deep spatio-temporal neural network, numerical weather prediction.

\maketitle

\section{Introduction}\label{sec1}
\begin{figure*}[!t]
\centerline{\includegraphics[width=\linewidth]{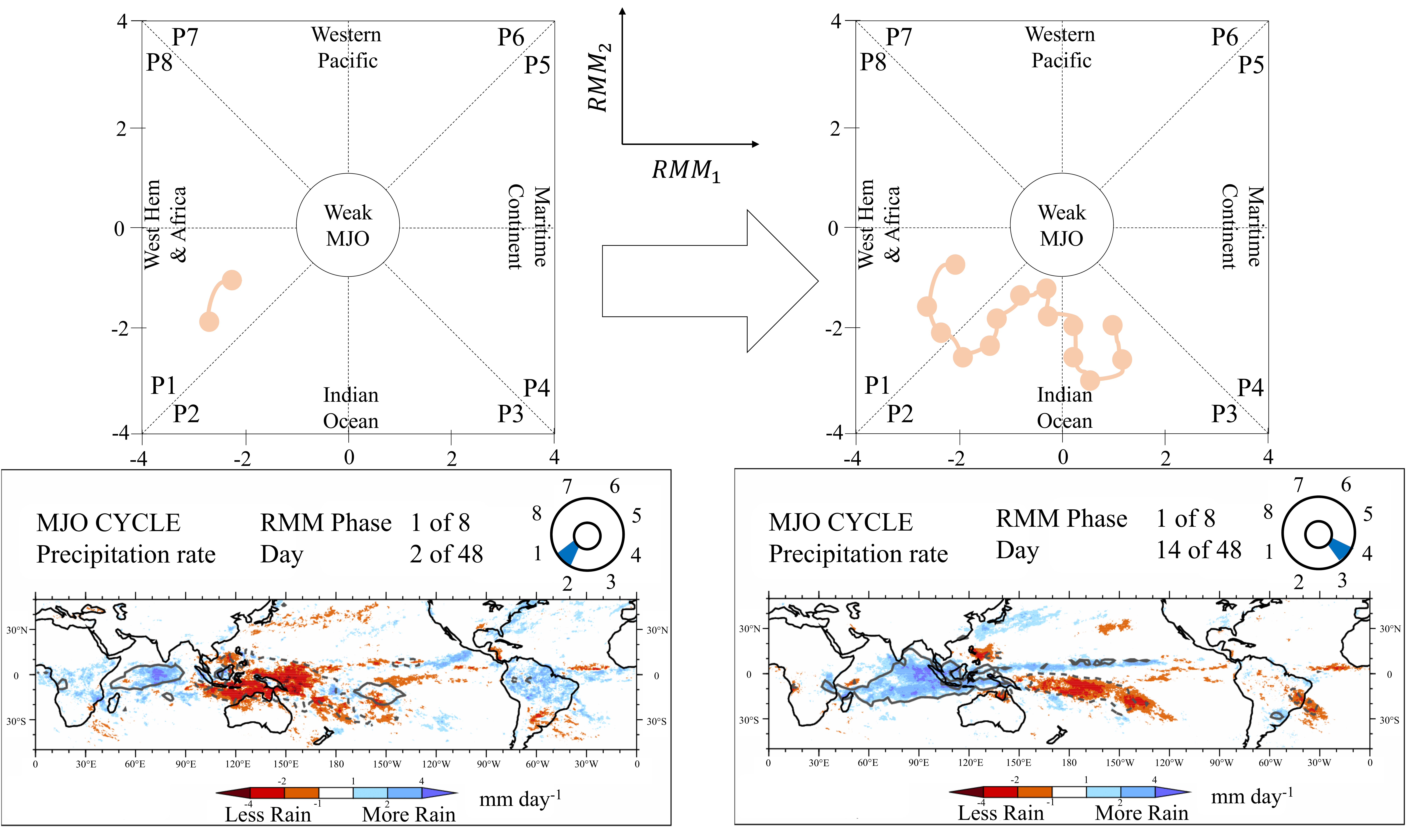}}
\caption{An example of MJO propagates eastward affecting precipitation, sourced from the online website (http://envam1.env.uea.ac.uk/mjo.html). And the RMM phase diagram shows the trajectory of MJO.}
\vspace{-0.5cm}
\label{fig1}
\end{figure*}

The Madden-Julian Oscillation (MJO)~\citep{b1,mjo1,mjo2}, which occurs each winter when tropical circulation and convection are coupled and migrate east, is one of the main sources of sub-seasonal to seasonal (S2S)~\citep{aaai, b3} weather predictions across the globe. A good S2S prediction (longer than two weeks but less than a season) offers decision-makers a valuable opportunity to learn about, for instance, changes in extreme event risk or opportunities to optimize resource management. The correct prediction of MJO is essential for S2S predictions to detect these extreme climate events, such as floods, high temperatures, and anomalous monsoons. Moreover, the MJO convection and circulation anomalies influence the weather and climate worldwide. To track MJO strength and location, researchers typically utilize MJO indexes known as Real-time Multivariate MJO (RMM)~\citep{b2}, as illustrated in Fig.~\ref{fig1}. Current MJO forecast methods include Numerical Weather Prediction (NWP)~\citep{b4,b5,nwp1,nwp2,nwp3,nwp4} and Artificial Neural Network (ANN)~\citep{b6,dp1,dp2,dp3,dp4,dp5,dp6,b48} based methods. The majority of NWP methods rely on Earth System Models, which convert geophysics principles into computational fluid dynamics, partial differential equations (PDEs), algebraic topology, and biological procedures solved numerically. And the MJO forecast based on NWP methods computes RMM indexes using NWP-predicted climate data. Currently, NWP methods provide the most accurate forecasts in MJO~\citep{mostacc,b6}, but simulating a 10-day forecast with a spatial resolution of $0.25^{\circ} \times 0.25^{\circ}$ will take hours of calculation with hundreds of supercomputer nodes~\citep{b7}. In addition, MJO forecasts based on NWP methods are sensitivity to seasonal variations, resulting in unstable predictions across different seasons. Most NWP methods provide better results in winter~\citep{unstable,b19}. For example, MJO forecasts from ECMWF, as tested in Section~\ref{5.4}, reach 34 days in winter but only 23 days in autumn.

Recent progress in deep learning~\citep{b8}, especially with ANNs, has made it possible for ANNs to explore and use climate prediction as a classification or regression task~\citep{b28,b52,b53}. There has been significant progress in recent years, with a paper~\citep{b28} published in Nature. For data-driven MJO forecasting with ANNs, neural networks are used to capture the relationship between input (climate data) and output (target data, i.e., RMM indexes). There are already many successful results based on ANN methods. As a recent example, auto-regressive recurrent neural network (AR-RNN)~\citep{b9} can forecast for less than 25 days with up to 300 days of RMM indexes data in 1-2 seconds on a single GPU. However, ANN methods still suffer from poor long-term prediction accuracy~\citep{aaai}. For the AR-RNN to accurately predict the MJO, it needs to be trained and inferred from nearly a year's worth of data, yet its highest prediction is still less accurate than the average of 28 days by the operational forecasts from ECMWF. This is due to the inequitable temporal interval between the input data used for training and inference and the output results. Consequently, predictions are likely to lose local features and be less accurate as a result. Meanwhile, AR-RNN prediction remains highly unstable due to seasonal influences. Currently, researchers continue to look for MJO forecasting methods that combine the benefits of both NWP and ANN.

This work focuses on improving ANN models' accuracy while ensuring stability and efficiency. By referencing~\citet{b28}, we want to create an ANN model for MJO forecasting based on atmospheric and oceanic circulation variable data. This is like an auto-regression problem based on spatio-temporal data in computer science. There are two major challenges recognized. Firstly, the observed sample of MJO events is deficient. The available data from the 5th generation of ECMWF reanalysis data (ERA5)~\citep{b10} beginning in 1950 shows only 60–90 days per year (December to February of the following year) of significant MJO events, implying that models lack training samples of MJO events. Due to this limitation, the neural network cannot capture information efficiently, resulting in continuously lower prediction accuracy than NWP. Secondly, traditional ANN methods are difficult to capture the spatio-temporal information of MJO effectively. It is aggravated by the lack of samples as well. In Fig.~\ref{fig1}, the RMM phase diagram shows the trajectory of MJO over time, which is nonlinearly distributed in both space and time. Moreover, climate data contain a large amount of information that is not relevant to MJO forecasts. These pieces of information cannot be actively removed by models, resulting in a more painful capture of MJO information.

In this paper, we propose DK-STN, a Domain Knowledge Embedded Spatio-Temporal Network model. It provides up to 28 days of stable seasonal prediction with high efficiency and significantly outperforms the latest ANN method (i.e., AR-RNN), while its accuracy is equivalent to the state-of-the-art NWP method (i.e., ECMWF). Technically, the DK-STN solves the aforementioned challenges in two aspects. \textbf{First}, meteorology reanalysis data provide less information about the MJO. In contrast, model data, which reforecast data using NWP methods, have superior accuracy. As a result, we employ a domain knowledge enhancement method based on model data to extend data diversity for network training. \textbf{Second}, we propose a domain knowledge processing method based on relevant meteorological knowledge to actively capture MJO information. In the meantime, we add it as a module (i.e., domain knowledge processing module, DKPM) to a spatio-temporal network (STN). This STN is designed to predict MJO reliably and efficiently. It has a spatial residual convolution module (SRCM) and a temporal adaptive attention module (TAAM) that fix network degradation~\citep{resnet} and an unbalanced ratio of MJO information in a long-term input, respectively. Compared to other typical spatio-temporal neural networks, our model does a better job of capturing the spatial and temporal information about MJO.

In summary, the contributions of this paper are summarized as follows:
\begin{itemize}
\item We propose DK-STN, a spatio-temporal neural network model embedded with climate domain knowledge. This model combines the strengths of NWP and ANN methods. Through training, DK-STN can process complex and evolving climate circulation data, generating accurate and stable  MJO forecasts with high efficiency. 
\item We reveal challenges impacting MJO forecast accuracy and address them in DK-STN using two methods: (i) a domain knowledge enhancement method-diversifying training data with the inclusion of numerical model hindcasts; and (ii) a domain knowledge processing method-incorporating spatio-temporal pre-processing methods inspired by the climate domain to capture MJO information.
\item Experiments show the remarkable predictive capabilities of DK-STN model.  It accurately forecasts MJO for up to 28 days, utilizing only 7 days of circulation variable data. Further, our model has an error of only 2-3 days between seasons and can infer the results in 2 seconds. While ensuring model efficiency and prediction stability, its accuracy outperforms other ANN methods, and is comparable as the state-of-the-art NWP method (i.e., ECMWF).
\end{itemize}

The remainder of this paper is organized as follows: Section~\ref{sec2} introduces related works. Section~\ref{sec3} presents the MJO forecasting problem formulation and analysis. Section~\ref{sec4} describes the proposed model in detail. This method will be compared with many other prediction models on a real meteorological dataset in Section~\ref{sec5}, and the results will be analyzed. Finally, in Section~\ref{sec6}, we will summarize the work of this study based on the experiment and discuss the future research path.

%\vspace{-10pt}

\section{Related Work}\label{sec2}

\subsection{Tracking the MJO}

To track MJO events, MJO indexes are widely used. Among them,~\citet{b2} introduced Real-Time Multivariate MJO indexes, $RMM 1$ and $RMM 2$, which are effective for tracing MJO's strength and position. RMM indexes represent the first and second components of combined empirical orthogonal functions (EOFs)\citep{eof}, derived from outgoing longwave radiation (OLR) and zonal wind at 200 and 850 hPa (U200, U850) over the global tropics. While other MJO indexes, such as OLR MJO index (OMI)\citep{b12}, Original OLR MJO index (OOMI)\citep{b13}, Real-Time OLR MJO index (ROMI)\citep{b14}, and Filtered OLR MJO index (FMO)~\citep{b15}, exist,  our paper employs RMM indexes.
%In 2004, Wheeler $et\ al.$~\cite{b2} introduced the Real-Time Multivariate MJO indexes, $RMM 1$ and $RMM 2$, which are the first and second principal components of the combined empirical orthogonal functions (EOFs)~\cite{eof} of the outgoing longwave radiationt (OLR) and the zonal wind at 200 and 850 hPa (U200, U850) averaged between global tropics, to track the strength and location of MJO. Then, many other MJO indexes emerged, including the OLR MJO index (OMI)~\cite{b12}, the Original OLR MJO index (OOMI)~\cite{b13}, the Real-Time OLR MJO index (ROMI)~\cite{b14}, and the Filtered OLR MJO index (FMO)~\cite{b15}, etc.

\begin{small}
\begin{flalign}
&\ RMM_{amp} = \sqrt{RMM_{1}^{2} + RMM_{2}^{2}}\label{eq1} &
\end{flalign}
\end{small}

To identify MJO occurrence based on strength, we utilize $RMM_{amp}$ from Eq.~\eqref{eq1}. A value above 1 indicates the presence of MJO. Additionally, by employing $RMM_1$ and $RMM_2$, we construct the spatial phase map (Fig.~\ref{fig1}), precisely locating the MJO position. The map delineates eight distinct phases, where phases 1 and 8 represent the Western Hemisphere and Africa. Phases 2 and 3 cover the Indian Ocean, while phases 4 and 5 encompass the Maritime Continent, and phases 6 and 7 include the Western Pacific. Each phase corresponds to a specific MJO region. Connecting daily RMM indexes points illustrates the MJO's activity trajectory, typically counterclockwise, reflecting eastward propagation.

\subsection{Predicting the MJO}

MJO forecasts include Numerical Weather Prediction (NWP) and Artificial Neural Network (ANN) methods. NWP, commonly employed in climate prediction, utilizes partial differential equations (PDEs) to simulate weather conditions over specific periods. MJO forecasts with NWP methods involve using the predicted data from NWP (e.g., OLR, U200, and U850) to calculate RMM indexes. In 2015, climate centers from 11 nations collaborated to create a Sub-seasonal to Seasonal (S2S) database~\citep{b16}, employing their respective NWP methods to obtain S2S model data. Two commonly used NWP methods among them are from the China Meteorological Administration (CMA)~\citep{b3} and the European Centre for Medium-Range Weather Forecast (ECMWF)~\citep{b3}. For convenience in subsequent narration, we refer to these methods as `CMA' and `ECMWF' respectively.

CMA accurately forecasts RMM indexes for approximately 21 days, while ECMWF achieves 27–28 days, currently the best~\citep{b7}. However, due to limited computing capabilities, solving increasingly complex PDEs over time is challenging for NWP methods, known for their resource-intensive nature. For instance, generating a 10-day forecast for a specific climate like OLR in ECMWF could take hours, requiring hundreds of supercomputer nodes. Additionally, MJO forecasts using NWP methods are sensitive to seasonal variations, resulting in less stable predictions across different seasons. Most NWP methods perform better in winter~\citep{unstable,b19}. ECMWF's MJO forecasts, as tested in Section~\ref{5.4}, extend to 34 days in winter but only 23 days in autumn.

Advancements in artificial intelligence, specifically ANN, offers new avenues for MJO forecast. ANN simplifies the complex PDE-solving process with an `end-to-end' model, approximating accurate predictions. Successful applications include a Fully Connected Neural Network (FNN) for a 17-day forecast~\citep{b19}, a neural network for MJO forecast bias correction with a 77-90$\%$ reduction in errors~\citep{b20}, and an AR-RNN predicting up to 25 days~\citep{b9}, albeit with a requirement of 300 days of RMM data as input, reflecting some inefficiency. However, the current accuracy of ANN methods still falls short of NWP methods.

\subsection{Spatio-temporal Neural Network}\label{DSTNN}

MJO forecasting, a complex task involving spatio-temporal dynamics, is well-suited for spatio-temporal neural networks. At present, numerous spatio-temporal neural network models have been applied in the field of climate, demonstrating remarkable predictive performance. For example, ConvLSTM~\citep{b21,b22} adopts convolution to model spatial dependencies in Long and Short-Term Memory Neural Network (LSTM)~\citep{b23} architecture. It converts hidden states to 3D tensors, replacing matrix multiplications with convolutions. This achieves excellent rainfall prediction. Similarly,~\citet{b24} implements convolutional networks within Gated Recurrent Unit Neural Network (GRU)~\citep{b25}, enabling high-resolution climate data modeling with maintained performance.
%MJO forecast is equivalent to a nonlinear auto-regressive problem with exogenous inputs based on spatio-temporal data since climate data contains both spatial and temporal properties. Researchers have developed deep spatio-temporal neural networks to address spatio-temporal data challenges, outperforming existing models in predicting climate. For example,~\citet{b21,b22} proposed a ConvLSTM for rainfall prediction, which converts the network's hidden states into a three-dimensional tensor and replaces matrix multiplication in Long and Short-Term Memory Neural Networks (LSTMs)~\citep{b23} with convolution, achieving powerful spatio-temporal prediction. Many comparable efforts, such as~\citet{b24}, embed convolutional networks within the variation of LSTM, namely Gated Recurrent Unit Neural Networks (GRUs)~\citep{b25}, enabling modeling of high-resolution (e.g., $0.25^{\circ} \times 0.25^{\circ}$) climate data. 

\begin{figure}[!t]
\centerline{\includegraphics[width=\linewidth]{fig2_lstm.pdf}}
\caption{The structure of a LSTM cell~\citep{b23}. ($f_{t}, i_{t}$ and $o_{t}$ represent forget gate, input gate and output gate respectively, $W_{*}$ and $ b_{*}$ represent weights and bias of each gate.)}
\vspace{-0.5cm}
\label{fig2}
\end{figure}

Observing these models, they comprise spatial models using Convolutional Neural Networks (CNNs)~\citep{b26,cnn,b27} and temporal models employing LSTM or GRU. Both LSTM and GRU are Recurrent Neural Networks (RNNs)~\citep{b31} adapted for long-term dependence, showing comparable effectiveness. In our model, we choose LSTM over GRU for the temporal model due to its superior information retention (see Section~\ref{M_T} for details). Fig.~\ref{fig2} illustrates an LSTM cell, including an additional cell $c_t$ for hidden information storage.

Tailored spatio-temporal neural networks are necessary for different climate predictions. For the MJO problem addressed in this paper, we conduct a detailed analysis due to its complexity compared to general climate problems.
%As can be observed, these spatio-temporal models are various combinations of spatial network models based on \textcolor{red}{Convolutional Neural Networks (CNNs)~\citep{b26,cnn,b27}} and temporal network models based on LSTM or GRU. Here, we discover that existing forecasts for climatic events including the El Niño-Southern Oscillation (ENSO)~\citep{b28} etc., are unable to use the state-of-the-art temporal model Transformer~\citep{b29,b30} due to a lack of training data. And both LSTM and GRU are the Recurrent Neural Network (RNN)~\citep{b31} changed to solve the long-term dependence problem~\citep{b32} and have comparable effects. In our model, we select LSTM rather than GRU as the foundation of the temporal network model since the former stores more information than the latter - see Section~\ref{M_T} for details. As depicted in Fig.~\ref{fig2} for one cell of LSTM, which has an additional cell $c_t$ to store hidden information. For different climate predictions, we need to design corresponding deep spatio-temporal neural networks. As for the MJO problem in this paper, we first analyzed it in greater detail since it is more complex than the general climate problem.

\section{MJO Forecasting Problem Formalization}\label{sec3}

In this paper, we aim to integrate the benefits of NWP and ANN methods. This is in order to create a neural network model capable of performing accurate, stable, and efficient MJO forecasting. In this regard, we formalize the MJO forecasting problem and examine the challenges found in its application. The goal of the MJO forecasting problem is to predict the RMM indexes for a future period based on relevant circulation variables across time. The problem modeling corresponding to neural networks requires the use of time series climate data as input. Then, we run them through a function to infer the output, i.e., the RMM indexes for a future period of time. We begin with the following definitions of input and output.

\begin{figure*}[!t]
\centerline{\includegraphics[width=\linewidth]{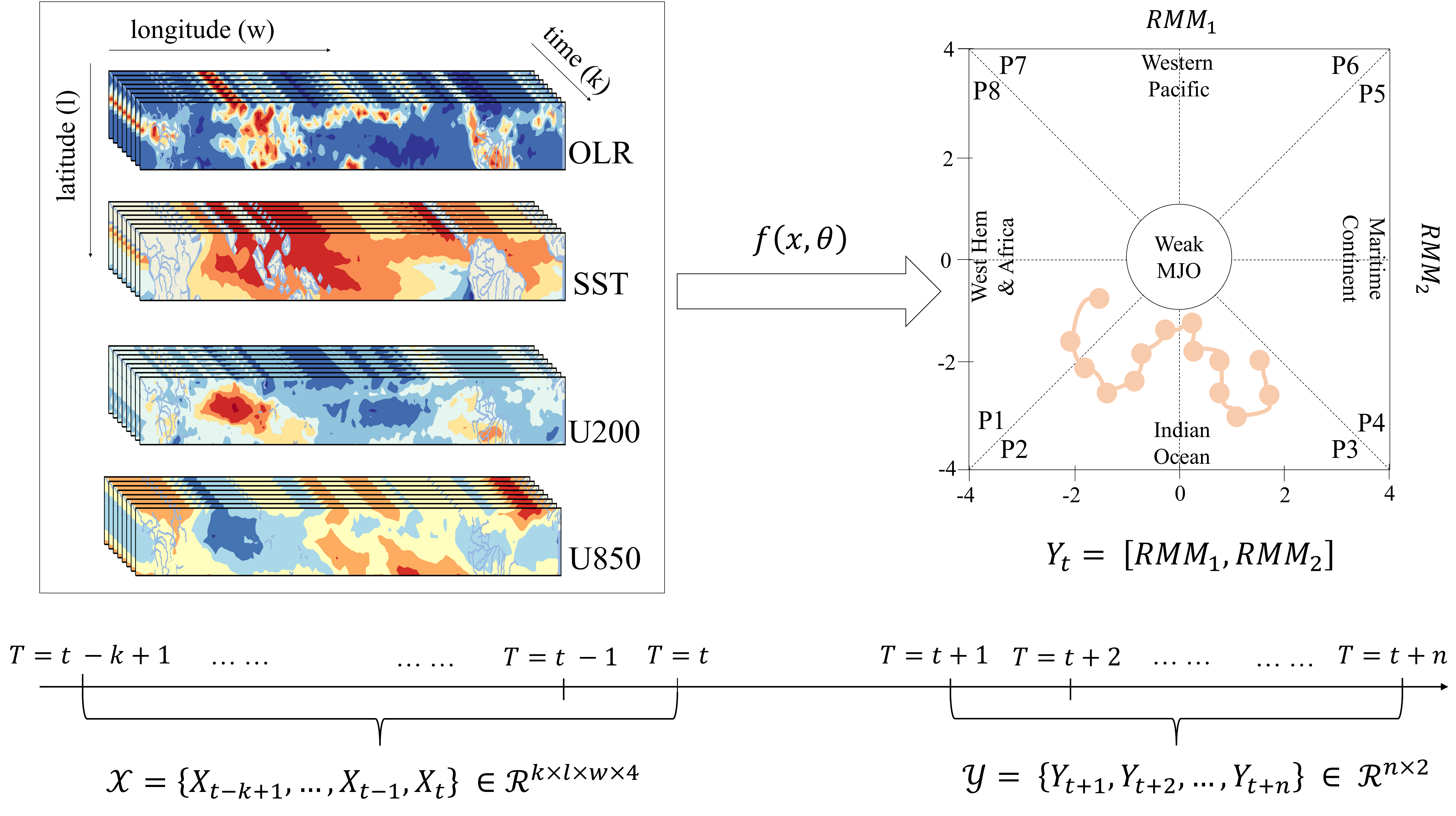}}
\caption{The MJO forecast problem formulation.}
\vspace{-0.5cm}
\label{fig3}
\end{figure*}

\noindent\textbf{Input. }Following prior work, we first divided the worldwide tropics $g$ ($0^{\circ}$-$360^{\circ}$, $15^{\circ}N$-$15^{\circ}S$) into a $l \times w$ (e.g., $2.5 \times 2.5$) grid image depending on longitude and latitude. Each grid is described as a region $r_{i,j}$ that is located in the $i_{th}$ row and $j_{th}$ column. Next, we define the related-climate impacting MJO throughout the time interval $t$ of all regions as an input data image donated as a matrix $X_t \in \mathcal{R}^{l \times w \times 4}$. Here, 4 indicates four atmospheric and oceanic circulation variables: OLR, U200, U850, and the Sea Surface Temperature (SST), which we learned are the primary elements affecting the MJO after consulting meteorological experts and examining relevant literature~\citep{b19}. Finally, given the latest time interval $t$ and historical time length $k$, we denote the input data image time-series as $\mathcal{X} = \left\{X_{t-k+1}, \ldots, X_{t-1}, X_t \right\} \in \mathcal{R}^{k \times l \times w \times 4}$, where $\mathcal{X}(t,i,j,*)$ denotes the OLR, U200, U850, and SST in region $r_{i,j}$ during $t$. In our paper, we choose the time interval $t=1$ day, and all the data are daily averages.

\noindent\textbf{Output. }As mentioned in Section~\ref{sec2}, we utilize RMM as the MJO index to track its strength and location. Each time interval $t$ (e.g., 1 day) has two RMM indexes, $RMM_1$ and $RMM_2$, which can be calculated using the matrix $X_t \in \mathcal{R}^{l \times w \times 3}$ (where 3 includes OLR, U200, and U850). So we define $RMM_1$ and $RMM_2$ as output MJO indexes during the time interval $t$, represented as a list $Y_t$. And, given the following time interval $t+1$ and the future time length $n$, we represent the output RMM indexes time-series as $\mathcal{Y} = \left\{Y_{t+1}, Y_{t+2}, \ldots, Y_{t+n} \right\} \in \mathcal{R}^{n \times 2}$, where $\mathcal{Y}(t,*)$ denotes the $RMM_1$ and $RMM_2$ in worldwide tropics during $t$.

As shown in Fig.~\ref{fig3}, we connect the input and output using a function $f$. And $f$ shows to be a neural network, which is frequently represented as $f(\mathcal{X}, \theta)$, where $\mathcal{X}$ is the input data and $\theta$ denotes the learnable parameters. A neural network frequently has multiple layers. Each of these layers includes a huge number of learnable parameters, which are initialized as white noise and optimized by back-propagating the prediction errors in the deep network. And we can acquire an optimal function $f$ that minimizes the error between anticipated and true values through extensive training and optimization as follows:

\begin{small}
\begin{flalign}
&\ \mathop{\min} \sum_{i=t+1}^{t+n} error(Y_{i}, \hat{Y}_{i}), \hat{\mathcal{Y}} = f(\mathcal{X}, \theta)\label{eq2} &
\end{flalign}
\end{small}

Here, $\hat{\mathcal{Y}} = \left\{\hat{Y}_{t+1}, \hat{Y}_{t+2}, \ldots, \hat{Y}_{t+n} \right\}$ is the predicted result while $\mathcal{Y}$ is the truth. $error$ can be measured using mean absolute error (MAE)~\citep{rmse}, root mean absolute error (RMSE)~\citep{rmse}, etc. According to~\citet{theory}, NWP methods are constrained by parameterization schemes and prediction results are affected by uncertain parameters. In contrast, ANN methods maximize likelihood probability in a data-driven manner without such restrictions. Therefore, given sufficient data, ANN-based MJO forecasting can get better results by finding the optimal function $f$. But in fact, we are still unable to match NWP performance. In this context, we explored the reasons that limit ANN forecasting MJO and discovered two major challenges listed below:

\begin{itemize}
\item \textbf{How to enhance data samples containing MJO events for model training?} It has always been a concern that climate data are scarce, which is why ANNs are constantly less accurate than NWP methods because neural networks are trained on a process of capturing relevant information from vast amounts of data and then aggregating it into an optimal function $f$. It has been proven that the more data there is, the more accurate the training will be as time passes. It is also worth pointing out that the data issue is significantly more acute in the case of MJO forecasting. Due to the incompleteness of observations, we often use the ERA5 reanalysis data for model training since the `reanalysis' involves reintegrating and optimizing diverse types and sources of observations using a state-of-the-art system~\citep{b10,b19}. There are around 26,000 samples from 1950 to the present, and only 6,000+ of these exhibit significant MJO events (i.e., $RMM_{amp} > 1$), which is insufficient for model training at 1-day time intervals and severely limits the network's feature capture. To improve the accuracy of ANN methods, we must feed enough data samples containing MJO events to the neural networks. However, the ERA5, with assured accuracy, only date back to 1950. As a result, our first challenge is to enhance the data over such a short period of time.

\item \textbf{How to capture spatio-temporal information about MJO with neural networks?} As indicated above, neural networks capture spatio-temporal information about related MJO events from the input data. Although many strong spatio-temporal neural networks have been used to capture the information, they are difficult to capture well since MJO occurs spatially in the worldwide tropics and temporally for 60-90 days from December to February of the following year, with a nonlinear distribution in both space and time, and with too small data samples. Additionally, we employ some conventional spatio-temporal neural networks to predict the MJO, including CNN+LSTM and ConvLSTM (see Section~\ref{sec5} for details); however, the effect of these networks is much lower than that of NWP. Moreover, we discover that the corresponding climate data (i.e., circulation variables including OLR, U200, U850, and SST) contain an excessive amount of information, such as inter-annual cycles, seasonal cycles, and ENSO signals. In contrast, the MJO information is more like the `noise' in data, yet neural networks cannot actively remove the majority and capture the `noise'. In addition, another challenge is to develop a suitable spatio-temporal neural network model in order to capture the information of MJO.
\end{itemize}

\section{Methodology}\label{sec4}

\begin{figure*}[!t]
\centerline{\includegraphics[width=\linewidth]{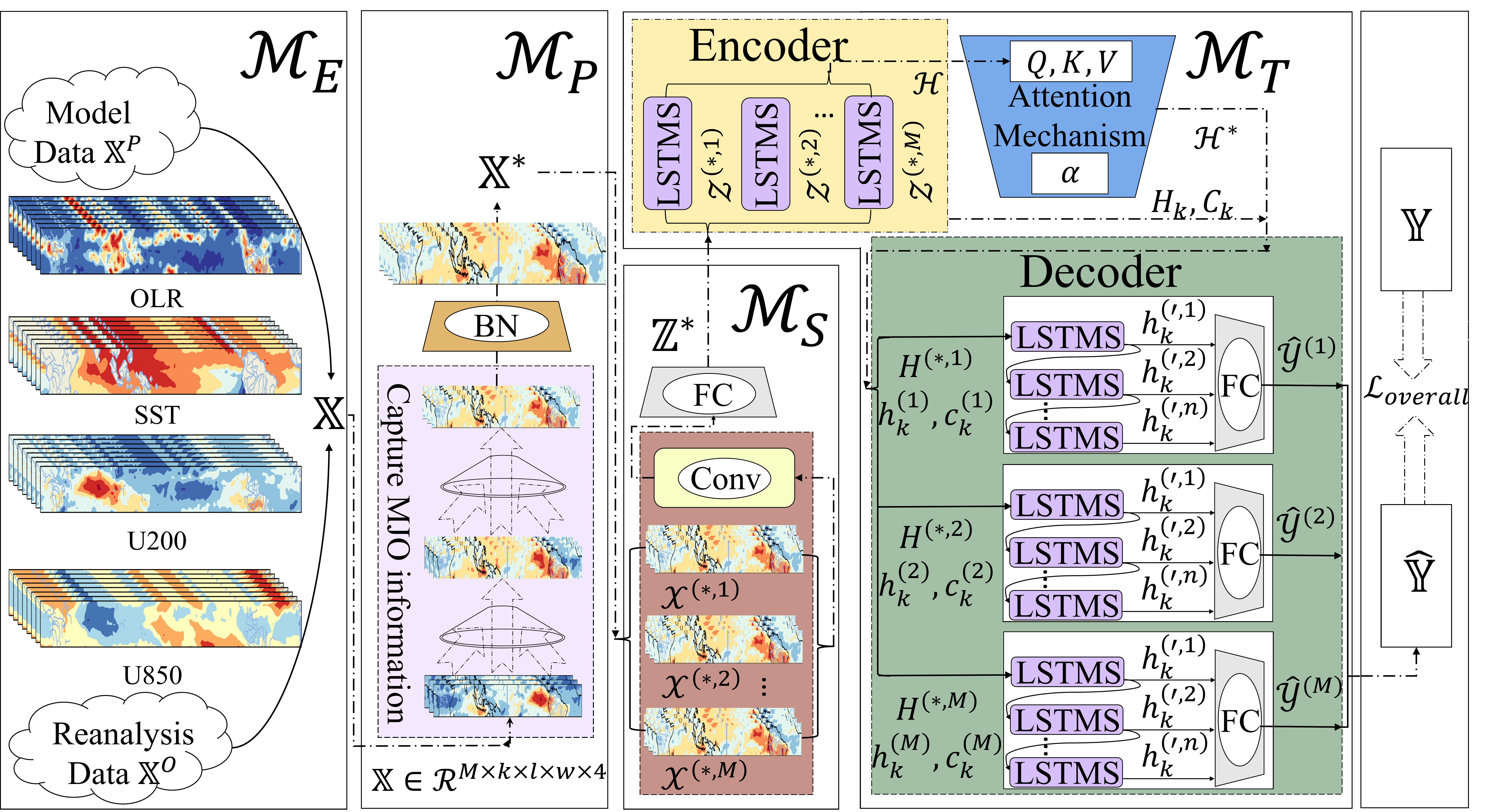}}
\caption{The proposed DK-STN Framework consists of a Domain Knowledge Enhancement Module $\mathcal{M}_E$, a Domain Knowledge Processing Module $\mathcal{M}_{P}$, a Spatial Residual Convolution Module $\mathcal{M}_{S}$, and a Temporal Adaptive Attention Module $\mathcal{M}_{T}$.}
\vspace{-0.5cm}
\label{fig4}    
\end{figure*}

\subsection{Framework Overview}
Our model, called Domain Knowledge Embedded Spatio-Temporal Network (DK-STN), is intended to address the above challenges by providing accurate, stable, and efficient MJO forecasting. It implements an ANN based method that successfully develops MJO forecasts using atmospheric and oceanic circulation variables and achieves the accuracy of the state-of-the-art NWP method while having excellent stability and efficiency. Fig.~\ref{fig4} depicts the DK-STN framework, which will be discussed in detail in this section.

With DK-STN, large quantities of climate data are used to train a spatio-temporal neural network that captures pertinent information about the MJO. In Section~\ref{M_E}, we describe a data enhancement method with domain knowledge and embed it as a Domain Knowledge Enhancement Module (DKEM) in DK-STN as network's data module, i.e., $\mathcal{M}_E$, while the most significant training part is divided into three major modules, i.e., $\mathcal{M}_P$, $\mathcal{M}_S$, and $\mathcal{M}_T$ (see Section~\ref{M_P}, Section~\ref{M_S}, and Section~\ref{M_T} respectively) to capture information about MJO actively. Moreover, since neural networks' training is intrinsically tied to the loss function, we recreate the overall loss function $\mathcal{L}_{overall}$ for MJO forecasting, which contains two losses $\mathcal{L}_1$ and $\mathcal{L}_2$ that correspond to $RMM_1$ and $RMM_2$ (see Section~\ref{loss}).

\subsection{Domain Knowledge Enhancement Module}\label{M_E}
We can only enhance data samples within the same time period, as stated in Section~\ref{sec3}, because we cannot improve data by expanding the time period (1950–). Currently, there are many data enhancement methods available for ANN, such as image duplication and rotation~\citep{b33}. It is pertinent to note, however, that these simple enhancement methods are inadequate to address the complex MJO problem and may result in network over-fitting. Furthermore, replication based solely on reanalysis data may result in unstable model predictions across seasons. This is because MJO events occur most frequently between December and February, when maximum prediction accuracy occurs. 

Our analysis of the necessary meteorological domain information shows that NWP methods have remarkably similar model data to truth because of their high prediction level of reforecasting data. Additionally, model data is more regular, which is due to the fact that NWP methods are composed of atmospheric dynamics equations that are relevant to the model. So model data are friendly to neural networks and may be simply fitted to their operations. Based on this, we propose a domain knowledge enhancement method that expands the dataset by adding series of model data. 

As shown in Fig.~\ref{fig4}, we integrate the method into $\mathcal{M}_E$ as the data input part of the neural network. We firstly choose $M/2$ time points to gather samples from the reanalysis data, with each sample including a continuous time length $k$, where the time interval is $t$ (e.g., 1 day), and obtain the input data set $\mathbb{X}^{O}=\left\{\mathcal{X}^{(O,1)}, \mathcal{X}^{(O,2)}, \ldots, \mathcal{X}^{(O,M/2)} \right\}\in \mathcal{R}^{\frac{M}{2} \times k \times l \times w \times 4}$. Each sample contains the OLR, U200, U850 and SST data over a period of $k$ days. For example, the  OLR, U200, U850 and SST data from Jan 1st, 1950 to Jan 7th, 1950 form one sample, i.e., $\mathcal{X}^{(O,1)} \in \mathcal{R}^{k \times l \times w \times 4}$. And for the model data, we identify the time points matching to the reanalysis data in different model data sets (e.g., CMA and ECMWF) and obtain the input model data set $\mathbb{X}^{P} \in \mathcal{R}^{\frac{M}{2} \times k \times l \times w \times 4} $.

Here, $M$ denotes the total number of training samples, with a 1:1 ratio between the reanalysis data and model data. To extend the dataset further, we use sliding windows of different lengths when selecting time points. Finally, we obtain the input data set $\mathbb{X} \in \mathcal{R}^{M \times k \times l \times w \times 4}$ as the input of our network via mixing $\mathbb{X}^{O}$ and $\mathbb{X}^{P}$.

\subsection{Domain Knowledge Processing Module}\label{M_P}

MJO information coexists with other information in climatic data and serves as `noise', making it difficult for neural networks to capture. As a result, we propose a domain knowledge processing method based on MJO index calculation, and we participate in training as $\mathcal{M_P}$ added to the neural network. As illustrated in Fig.~\ref{fig4}, $\mathcal{M_P}$ consists primarily of a captured MJO information part and a batch normalization layer.

\noindent\textbf{Capture MJO information.} Referring to~\citet{b2}, inter-annual cycles, seasonal cycles, and ENSO signals occupy most of the climate data, resulting in the MJO information being hidden. As part of this subsection, we sequentially process the data by combining relevant domain knowledge to extract information regarding MJO. For inter-annual and seasonal cycles, we use harmonic analysis~\citep{b2,b19} to remove the 0-3 waves from the corresponding climate for each data $\mathcal{X}(t,i,j,*)$ in set $\mathbb{X}$. Specifically, the daily data $\mathcal{X}(t)$ for the three circulation variables (i.e., OLR, U200, and U850) are subjected to the fourier expansion following Eq.\eqref{eq3}. The 0-3 waves of the daily climate states for each variable are then computed using Eq.\eqref{eq4}. Additionally, for the SST data, the land values masked as -32767 are all set to 0, without applying the harmonic analysis and accumulation. The processed SST data is concatenated with the OLR, U200 and U850 data processed by Eqs.~\eqref{eq3},~\eqref{eq4},~\eqref{eq5}, and~\eqref{eq6} as the input for the \textbf{Batch Normalization Layer}.

\begin{small}
\begin{flalign}
& \mathcal{X}(t) = a_0 + \sum_{n=1}^{\infty}(a_n\cos{(n \omega t)} + b_n\sin{(n \omega t)})\label{eq3} \\
& \Bar{\mathcal{X}}^{3}(t) = a_0 + \sum_{n=1}^{3}(a_n\cos{(n \omega t)} + b_n\sin{(n \omega t)})\label{eq4} &
\end{flalign}
\vspace{-0.2cm}
\end{small}

Here, $a_0$ represents the average of $\mathcal{X}(t)$, and $a_n/b_n$ represent the cos/sin function coefficients of the $n_{th}$ wave. Additionally, $n$ denotes the harmonic number, and $\omega$ denotes the circular frequency. Finally, we remove the 0-3 waves $\Bar{\mathcal{X}}^{3}(t)$ (i.e., inter-annual and seasonal cycles) from each data set $\mathcal{X}(t)$ to produce the updated data set $\mathcal{X}^{'}(t)$, as given in Eq.~\eqref{eq5}.

\begin{small}
\begin{flalign}
&\ \mathcal{X}^{'}(t) = \mathcal{X}(t) - \bar{\mathcal{X}}^{3}(t)\label{eq5} \\
&\ \mathcal{X}^{''}(t) = \mathcal{X}^{'}(t) - (\sum_{i=1}^{120}\mathcal{X}^{'}(t-i))/120\label{eq6} &
\end{flalign}
\end{small}

As for ENSO signals, we refer to our climate domain knowledge and understand that they are most often based on an average of the first 120 days of circulation variable data~\citep{b2,b35}. As a result, in Eq.~\eqref{eq6}, we calculate and subtract correspondingly to obtain the captured information $\mathcal{X}^{''}(t)$.

\noindent\textbf{Batch Normalization Layer.} We know that training neural networks is tough, and as the number of layers in a neural network increases, the distribution of input data changes, making training difficult. Batch Normalization (BN)~\citep{b36} is a technique that was developed to address this issue. By using this method, the input to a neural network can be normalized, as well as the mean and standard deviation of the input adjusted. In this manner, we will be able to fix the data distribution, which will significantly enhance the performance and stability of the neural network. 

In this subsection, we put the BN layer into our network's processing module $\mathcal{M}_P$, and the $M$ data $\mathcal{X}^{''}(t)$ just obtained from processing are separated into batches and normalized independently, with the $mean$ and $std$ of each batch contributed to the training as neural network parameters. 

According to $\mathcal{M}_P$, we obtain normalized data while containing mainly MJO information: $\mathbb{X}^{*} \in \mathcal{R}^{M \times k \times l \times w \times 4}$

\subsection{Spatial Residual Convolution Module}\label{M_S}

As described in Section~\ref{DSTNN}, we find that spatio-temporal neural networks consist of spatial and temporal networks combined in a variety of ways, with spatial networks predominantly used as convolutional networks. On this foundation, we design the spatial residual convolution module (SRCM) shown in Fig.~\ref{fig4}, which includes a convolutional layer and a fully connected layer.

\noindent\textbf{Residual Convolutional Layer.} We apply a residual convolutional neural network (ResNet)~\citep{b37,b38}, like in most of the previous studies, to capture spatial information since the residual structure enables improved image data retention (we see the input as images). Specifically, we get $M\times k$ data $X^{*} \in \mathcal{R}^{l \times w \times 4}$ from set $\mathbb{X}^{*}$ without considering the time-series firstly, where $X^{*}(i,j,c)$ denotes the circulation variable data in region $r_{i,j}$. And $i,j,c$ just respectively correspond to the length, width and channel of images which use convolution for training usually. Next, each $X^{*}$ is respectively fed into residual convolution neural network as follows. Here, $\theta_{\text{layer}}$ denotes all parameters involved in the training of each convolutional layer. In our paper, we configure a total of 7 layers.

\begin{equation}
Z =  
\begin{cases}
\left.
\begin{aligned}
& conv(X^*, \theta_{\text{layer}}) & \text{if } \text{layer} = 1, \\
& relu(Z_{(\text{layer}-1)} + conv(Z_{(\text{layer}-1)}, \theta_{\text{layer}})) & \text{otherwise}.  
\end{aligned}
\right.
\end{cases}
\label{eq7}
\end{equation}

According to Eq.~\eqref{eq7}, we can obtain the initial spatial information set of $\mathbb{Z} = \left\{Z_{t-k+1}^{(1)}, \ldots, Z_{t}^{(1)}; Z_{t-k+1}^{(2)}, \ldots, Z_{t}^{(2)}; \right.$  $\left. Z_{t-k+1}^{(M)}, \ldots, Z_{t}^{(M)}\right\} \in \mathcal{R}^{M \times k \times L \times W \times C}$, where $L\times W$ is the dimension of the two-dimensional matrix after paving, and $C$ is the output feature dimension.

\noindent\textbf{Fully Connected Layer.} To input the temporal module, we use the fully connected (FC)~\citep{b39} layer to expand the set $\mathbb{Z}$ by time-series. A flattening function is used on each of $M \times k$ samples in the FC to obtain all spatial information.  As a result, we obtain an updated data set $\mathbb{Z}^{*} \in \mathcal{R}^{M \times k \times \hat{C}}$ of information containing the processing space, where $\hat{C} = L \times W \times C$.
%\left\{\mathcal{Z}^{(*,1)}, \mathcal{Z}^{(*,2)}, \ldots, \mathcal{Z}^{(*,M)} \right\} =
\subsection{Temporal Attentive Adaptation Module}\label{M_T}

A temporal module is required to capture the temporal characteristics of the data following SRCM. The most appropriate current temporal model is Transformer~\citep{b29}, as described in Section~\ref{DSTNN}; however, it is not ideal for learning from small samples. As a result, we design an Seq2Seq structure~\citep{b40,b41} with an attention mechanism~\citep{b42} as the temporal attentive adaptation module (TAAM), from which Transformer evolves, but it can capture temporal information effectively with small samples.

As shown in Fig.~\ref{fig4}, TAAM is designed for time-aware knowledge in different regions as MJO is also changed in the temporal dimension. We first use the Encoder in Seq2Seq to capture the domain weights, which represent the impact of temporal features in the global tropics. Then, an attention mechanism is employed to obtain different contributions of temporal information that represent each time interval. Finally, we use the Decoder in Seq2Seq to get prediction results, $RMM_1$ and $RMM_2$.

\begin{figure}[!t]
\centerline{\includegraphics[width=\linewidth]{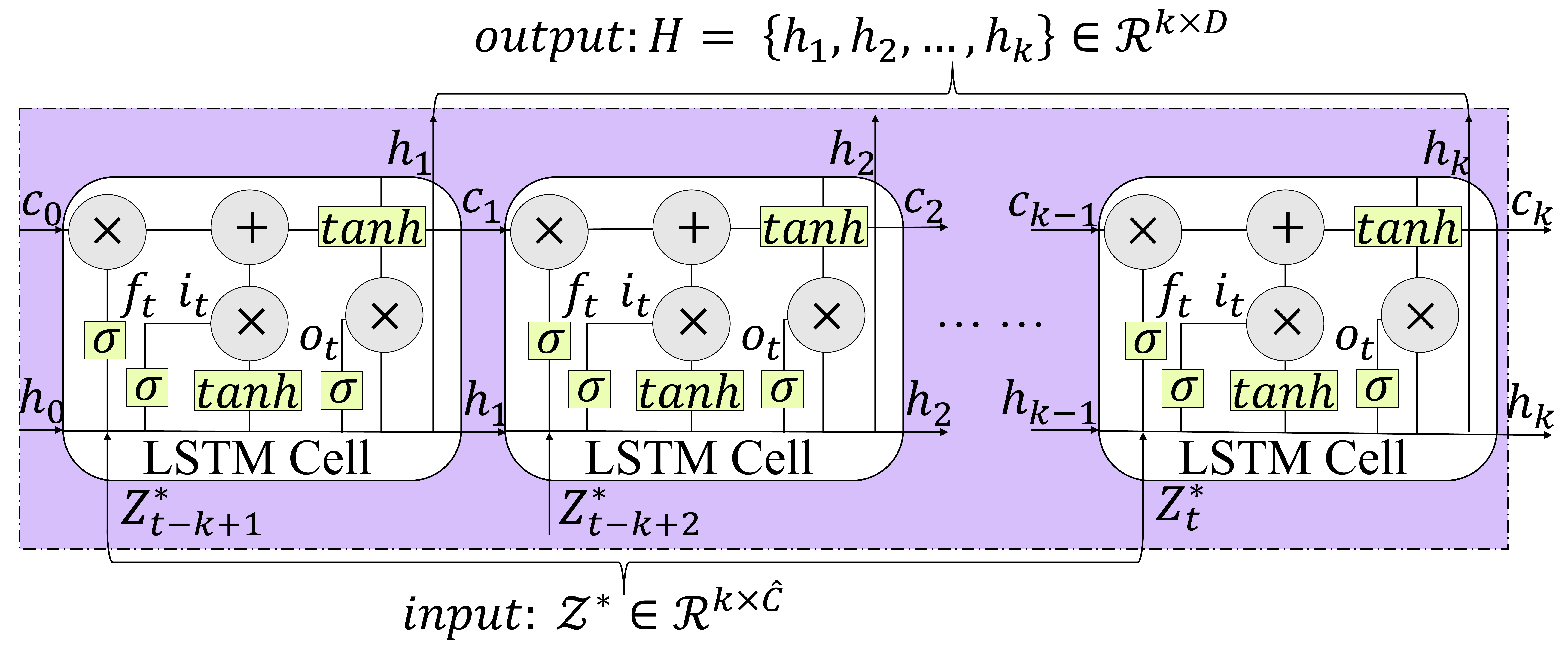}}
\caption{The structure of LSTMS.}
\vspace{-0.5cm}
\label{fig5}    
\end{figure}

\noindent\textbf{Encoder.} In the Encoder, we select the LSTM for local temporal information embedding because the LSTM can capture the correlation in time series and store more temporal information than an RNN or GRU due to the extra $c_{t}$. As seen, we feed $\mathbb{Z}^{*}$ to $M$ Long and Short-Term Memory Sequences (LSTMSs). As shown in Fig.~\ref{fig5}, the time sequence of each LSTMS is $k$, i.e., each LSTMS is connected in series by $k$ LSTM cells. (Fig.~\ref{fig2} depicts a LSTM cell's structure, and each one generates two states, $h_{t}$ and $c_{t}$, which respectively denote $D$ temporal features and cell states in the global tropics during time interval $t$.) For each LSTMS, we can obtain an output set $H \in \mathcal{R}^{k \times D}$ and the final states $h_{k}$ ,$c_{k}$ with $\mathcal{Z}^{*}$.

After connecting the outputs of each LSTMS, we compute the initial temporal information set of $\mathbb{Z}^{*}$, including $\mathcal{H} \in \mathcal{R}^{M \times k \times D}$ and the final state sets, i.e., $H_{k} \in \mathcal{R}^{M \times D}$ and $C_{k}$.

\noindent\textbf{Attention Mechanism.} Referring to~\citet{b43}, not all temporal features within a time range $k$ will contribute equally owing to temporal dependence, and therefore, temporal correlation would be critical. To address this, as shown in Fig.~\ref{fig4}, $\mathcal{M}_T$ introduces an attention mechanism between the encoder and decoder to consider different contributions.

To be more specific, we design an attention weight based on the scaled dot product attention method~\citep{b42,b43,b44} to capture those critical temporal information in the time range $k$. The weight $\alpha$ of each set $H$ is defined as:

\begin{small}
\begin{flalign}
\begin{split}
& Q = H * W^{Q} , K = H * W^{K}, V = H * W^{V}\\ 
& \alpha = softmax(\frac{Q * K^{T}}{\sqrt{d_{k}}})\label{eq8} 
\end{split}&
\end{flalign}
\end{small}

where $Q,K,V \in \mathcal{R}^{k \times D}$ represent query vector, key vector and value vector, respectively. $W^{Q}, W^{K}, W^{V} \in \mathcal{R}^{D \times D}$ are random initialization parameter matrices, $\alpha \in \mathcal{R}^{k \times k}$ is the weight matrix. $d_{k}$ is the dimension of $Q$ or $K$. Then, we get a brand-new temporal information set, i.e., $H^{*}$ by performing a weighted sum method of the value and attention weight $\alpha$ as shown in Eq.~\eqref{eq9}.

\begin{small}
\begin{flalign}
&\ H^{*} = \alpha * V\label{eq9} &
\end{flalign}
\end{small}

By connecting $M$ temporal information set $H^{*}$, which is distinguished by different contributions within the time range $k$, we get the set $\mathcal{H}^{*}$ as inputs to the Decoder.

\noindent\textbf{Decoder.} As depicted in Section~\ref{sec3}, MJO forecast is a multi-step regression task, and thus we design a decoder in $\mathcal{M}_T$ to get prediction results. We continue to use LSTMS for temporal information decoding and use $M \times n$ LSTMSs as inputs (including the temporal feature set $\mathcal{H}^{*}$ and final state sets $H_k, C_k$) because our goal is to predict results over the next $n$ time range. The time sequence of each LSTMS is $k$ as well.

As observed in Fig.~\ref{fig4}, we divide the decoder into $M$ parallel regions and each of them contains $n$ tandem LSTMSs. We feed $H^{*} \in \mathcal{R}^{k \times D}$ obtained through the attention mechanism and $h_{k}, c_{k}$ from the encoder to the first LSTMS in each region and predict the outputs (i.e., $H^{'}$, $h_{k}^{'}$, and $c_{k}^{'}$) of the next moment in the future. Then these outputs are used as inputs for the second LSTMS, $\cdots$. By looping, we get the states $H_{k}^{'} \in \mathcal{R}^{n \times D}$ from $n$ LSTMSs.

Next, we further compress each $h_{k}^{'} \in \mathcal{R}^{D}$ in the $H_{k}^{'}$ via a linear function which is as the FC layer to get $n$ predicted results, and each contains 2 RMM indexes. The predicted results can be denoted as $\mathcal{\hat{Y}}$.

Finally, after connecting the predicted results of each region, we get the final set of predicted results, i.e., $\mathbb{\hat{Y}} = \left\{ \mathcal{\hat{Y}}^{(1)}, \mathcal{\hat{Y}}^{(2)}, \ldots, \mathcal{\hat{Y}}^{(M)}\right\} \in \mathcal{R}^{M \times n \times 2}$.

\subsection{Overall Loss Function}\label{loss}
Referring to ~\citet{b45}, the Mean Squared Error Loss (MSELoss) function, which is widely used in various spatio-temporal prediction studies, we reconstruct the overall loss function for $RMM_1$ and $RMM_2$ to train the DK-STN. The loss function $\mathcal{L}_{overall}$ is as follows:

\begin{small}
\begin{flalign}
    \label{eq10}
    \begin{split}
        \mathcal{L}_{overall} & = \beta \mathcal{L}_{1} + \gamma \mathcal{L}_{2}\\
        & = \beta(\frac{1}{M} \sum_{i=1}^{M} (\frac{1}{n} \sum_{j=1}^{n} \left\| \mathbb{\hat{Y}}(i,j,1) - \mathbb{Y}(i,j,1)\right\|^{2}))\\
        & + \gamma(\frac{1}{M} \sum_{i=1}^{M} (\frac{1}{n} \sum_{j=1}^{n} \left\| \mathbb{\hat{Y}}(i,j,2) - \mathbb{Y}(i,j,2)\right\|^{2}))
    \end{split} &
\end{flalign}
\end{small}

Here, $\mathcal{L}_1$ and $\mathcal{L}_2$ denote the mean squared losses of $RMM_1$ and $RMM_2$ respectively and $\beta$ and $\gamma$ control the weight of the two RMM indexes. Furthermore, $\mathbb{\hat{Y}}$ is the predicted result set, whereas $\mathbb{Y}$ is its corresponding truth, or called the label for the model, where $\mathbb{\hat{Y}}(i,j,1)/\mathbb{Y}(i,j,1)$ denotes the $RMM_1$ in $i$-th data sample during $j$-th time interval and $\mathbb{\hat{Y}}(i,j,2)/\mathbb{Y}(i,j,2)$ denotes the $RMM_2$. In our paper, we set $\beta$ and $\gamma$ as 0.5.

Finally, we incorporate the above process into Algorithm.~\ref{alg}, elucidating our model training procedure. The input is the merged dataset $\mathbb{X}$, with the epoch and learning rate specified as detailed in Section~\ref{e_s}. Model parameters involved in training are initialized.

Subsequently, the model undergoes multiple epochs of iterations. In each iteration, the loss between predicted and true labels is computed using Eq.~\eqref{eq10}. Gradient descent is then employed to update parameters, leveraging the derivative of the loss with respect to parameters. This iterative process allows us to discover optimized parameters through multiple iterations. With the optimized parameters, we obtain our trained model, capable of forecasting MJO.

\begin{algorithm}
  \caption{DK-STN}\label{alg}
  \begin{algorithmic}[1]
    \Procedure{DK-STN}{$\mathbb{X}$, \textit{epochs}, \textit{learning\_rate}}
      \State \textbf{Input:} $\mathbb{X}$ \Comment{Dataset $\mathbb{X} = \mathbb{X}^O \cup \mathbb{X}^P$ containing $M$ samples.} 
      
      \State Initialize parameters: $\theta$

      \For{\textit{epoch} $\gets 1$ to \textit{epochs}}
        \State $\mathbb{X} \gets \mathbb{X}(t) - \Bar{\mathbb{X}}^{3}(t)$ \Comment{Eqs.~\eqref{eq3},~\eqref{eq4} and~\eqref{eq5}}
        \State $\mathbb{X}^{''} \gets (\mathbb{X}^{'} - \text{avg}(\mathbb{X}^{'}[:120]))$ \Comment{Eq.~\eqref{eq6}}
        \State $\mathbb{X}^{''} \gets \text{Batch Normalization Layer}(\mathbb{X}^{''})$
        \State $\mathbb{Z} \gets \text{Residual Convolutional Layer}(\mathbb{X}^{''})$ \Comment{Eq.~\eqref{eq7}}
        \State $\mathbb{Z}^{*} \gets \text{Fully Connected Layer}(\mathbb{Z})$

        \State $\mathcal{H}, H_{k}, C_{k} \gets \text{Encoder}(\mathbb{Z}^{*})$
        \State $\mathcal{H}^{*} \gets \text{Attention Mechanism}(\mathcal{H})$ \Comment{Eqs.~\eqref{eq8} and~\eqref{eq9}}
        \State $\mathbb{\hat{Y}} \gets \text{Decoder}(\mathcal{H}^{*}, H_{k}, C_{k})$ \Comment{Prediction}

        \State $loss(\theta) = \mathcal{L}_{overall}(\mathbb{\hat{Y}}, \mathbb{Y})$ \Comment{Eq.~\eqref{eq10}}
        \State $\frac{\partial loss(\theta)}{\partial \theta} \gets \text{Compute Gradient}$ \Comment{Gradient computation}

        \State $\theta \gets \theta - \textit{learning\_rate} \times \frac{\partial loss(\theta)}{\partial \theta}$ \Comment{Gradient descent update}
      \EndFor

      \State \textbf{Return} Optimal Parameters $\theta$
    \EndProcedure
  \end{algorithmic}
\end{algorithm}
\vspace{-0.4cm}
\section{Experiment}\label{sec5}

\subsection{Experiment settings}\label{e_s}

\noindent\textbf{Datasets.} We conducted experiments on the datasets covering the entire equatorial region. Table~\ref{tab1} summarizes the statistics of the chosen data. Initially, we selected four types of circulation variables, namely OLR (\si{W/m^2}), U200 (\si{m/s}), U850 (\si{m/s}), and SST (\si{K}), from the ERA5 data and model data (CMA and ECMWF). Subsequently, we screened and divided the global tropical regions (i.e., $15^{\circ}$N to $15^{\circ}$S latitudes and $0^{\circ}$ to $360^{\circ}$ longitudes) based on a resolution of $2.5^{\circ} \times 2.5^{\circ}$. Following this, daily data were selected from January 1, 1950, 2006, and 2001 to December 31, 2021. To further expand the data, we constructed different sliding windows with lengths of 2, 7, and 1 days for different data sources, respectively. Finally, we utilized data before December 31, 2019, for training and data from January 1, 2020, to December 31, 2020, for validation. The remaining ERA5 data, spanning January 1, 2021, to December 31, 2021, was used for testing.
%We conducted experiments on the datasets we created, which cover the whole equatorial region. Table~\ref{tab1} summarizes the statistics of the data that we chose. Specifically, we first selected four types of circulation variables, \textcolor{red}{OLR (\si{W/m^2}), U200 (\si{m/s}), U850 (\si{m/s}), and SST (\si{K})}, from the ERA5 data and the model data (\textcolor{red}{CMA and ECMWF}). Then, the global tropical regions (i.e., $15^{\circ}$N to $15^{\circ}$S latitudes and $0^{\circ}$ to $360^{\circ}$ longitudes) are screened and divided according to the resolution of $2.5^{\circ} \times 2.5^{\circ}$. \textcolor{red}{Next, we selected daily data from January 1, 1950, 2006, and 2001 to December 31, 2021, in terms of time. Simultaneously, to expand the amount of data further, we constructed different sliding windows with lengths of 2, 7, and 1 days for different sources of data, respectively.} Finally, we use data before December 31, 2019 for training and data from January 1, 2020, to December 31, 2020, for validating. The rest of the ERA5 data, which is from January 1, 2021, to December 31, 2021, is used for testing.

\noindent\textbf{Evaluation Metrics.} Referring to relevant meteorological studies~\citep{b6,b46}, we adopt two commonly used metrics to evaluate the prediction skill of MJO: the bivariate correlation coefficient (\textbf{COR}) and the root mean square error (\textbf{RMSE}). The metrics are defined as follows: 

\begin{small}
\vspace{-0.5cm}
\begin{flalign}
\begin{split}
&{\rm COR}(j) = \\
&\frac{\sum_{i=1}^M [\mathbb{\hat{Y}}(i,j,1) \times \mathbb{Y}(i,j,1) + \mathbb{\hat{Y}}(i,j,2) \times \mathbb{Y}(i,j,2)]}{\sqrt{\sum_{i=1}^M [\mathbb{\hat{Y}}^2(i,j,1) + \mathbb{\hat{Y}}^2(i,j,2)]}\sqrt{\sum_{i=1}^M [\mathbb{Y}^2(i,j,1) + \mathbb{Y}^2(i,j,2)]}}\\\label{eq11}
\end{split}\\
\begin{split}
&{\rm RMSE}(j) = \\
&\sqrt{\frac{1}{M}\sum_{i=1}^M [\lvert \mathbb{\hat{Y}}(i,j,1) - \mathbb{Y}(i,j,1) \rvert ^2 + \lvert \mathbb{\hat{Y}}(i,j,2) - \mathbb{Y}(i,j,2) \rvert ^2]}\\\label{eq12}
\end{split}&
\end{flalign}
\end{small}

Additionally, to demonstrate the accuracy of various methods in predicting both the amplitude and phase of the MJO, we introduce two new metrics: the average amplitude error (\textbf{$\overline{\rm AE}$}) and the average phase error (\textbf{$\overline{\rm PE}$}). The metrics are defined as follows:

\begin{small}
\begin{flalign}
\begin{split}
&{\overline{\rm AE}}(j) = \\
&\frac{1}{M}\sum_{i=1}^M \left(\sqrt{\mathbb{\hat{Y}}^2(i,j,1)+\mathbb{\hat{Y}}^2(i,j,2)} - \sqrt{\vphantom{\mathbb{\hat{Y}}^2(i,j,1)+\mathbb{\hat{Y}}^2(i,j,2)}\mathbb{Y}^2(i,j,1)+\mathbb{Y}^2(i,j,2)} \right)\\\label{eq13}
\end{split}\\
\begin{split}
&{\overline{\rm PE}}(j) = \frac{1}{M}\sum_{i=1}^M \left[ \arctan\left(\frac{\mathbb{\hat{Y}}(i,j,2)}{\mathbb{\hat{Y}}(i,j,1)}\right) - \arctan\left(\frac{\mathbb{Y}(i,j,2)}{\mathbb{Y}(i,j,1)}\right)\right]\\\label{eq14}
\end{split}&
\end{flalign}
\end{small}

The descriptions of variables in Eqs.~\eqref{eq11},~\eqref{eq12},~\eqref{eq13}, and~\eqref{eq14} can be found in Section~\ref{M_T}. ${\rm COR}(j)$ indicates the co-occurrence strength between the prediction and the truth during the $j$-th time interval within the subsequent time range $n$, while ${\rm RMSE}(j)$ conducts a term-by-term comparison of the actual difference, with the time interval being one day. Skill thresholds commonly used are ${\rm COR} = 0.5$ and ${\rm RMSE} = 1.4$: the prediction skill is determined when ${\rm COR}(j)$ falls below \textbf{0.5} and ${\rm RMSE}(j)$ exceeds \textbf{1.4}, corresponding to the forecast lead time of $\textbf{j}$ days. Additionally, $\overline{\rm AE}$ and $\overline{\rm PE}$ assess errors in amplitude and phase forecasts relative to their true values. A smaller $\overline{\rm AE}$ closer to 0 indicates lower amplitude error, meaning a better match between predicted and true amplitudes. Similarly, a lower $\overline{\rm PE}$ value signifies a smaller variation in phase shifts, suggesting better MJO phase prediction.
%\textcolor{red}{The relevant descriptions of variables in Eqs.~\eqref{eq11},~\eqref{eq12},~\eqref{eq13}, and~\eqref{eq14} are described in Section~\ref{M_T}.} And ${\rm COR}(j)$ expresses the strength of co-occurrence between the prediction and the truth during the $j$-th time interval in the following time range $n$, whereas ${\rm RMSE}(j)$ does a term-by-term comparison of the actual difference. (Here, the time interval is one day.) The values ${\rm COR} = 0.5$ and ${\rm RMSE} = 1.4$ are commonly used as skill thresholds: the prediction skill refers to the time $\textbf{j}$ when the ${\rm COR}(j)$ falls less than \textbf{0.5} and ${\rm RMSE}(j)$ grows greater than \textbf{1.4}, i.e., the forecast lead time is $\textbf{j}$ days. \textcolor{red}{Furthermore, $\overline{\rm AE}$ and $\overline{\rm PE}$ are used to assess the errors in amplitude and phase forecasts relative to their respective true values, respectively. Smaller $\overline{\rm AE}$ closer to 0 indicates lower amplitude error and better match between predicted and true amplitudes. Similarly, lower $\overline{\rm PE}$ value means smaller variation of phase shifts, suggesting better MJO phase prediction.}

\noindent\textbf{Model Parameters.} We begin by dividing the global tropics into grid maps and segmenting the time dimension into intervals (see Table~\ref{tab1}). Subsequently, we choose reforecasting data, where each period consists of 7 time intervals (i.e., 7 days). RMM indexes are computed for each day as labels for supervised learning based on~\citep{b2}. For $\mathcal{M}_S$, convolution kernel sizes are set to $7 \times 7$ and $3 \times 3$. Regarding $\mathcal{M}_T$, the length of LSTM and hidden feature size in the Encoder are set to 7 and 256, respectively, while in the Decoder, they are set to 35 and 256 (i.e., $k=7$, $n=35$). We choose $n=35$ days for a comprehensive comparison, considering ECMWF's high skill for 34 days in winter (details in Fig.~\ref{fig8} and Fig.~\ref{fig9}). Moreover, reducing to $n=30$ provides minimal performance gain. Hence, we opt for $n=35$ to facilitate model comparison. Regarding model training, the epoch size, weight decay, and learning rate are set to 25, 0.001, and $1 \times e^{-4}$, respectively. We use Adam as our optimizer, and it includes parameters for weight decay and learning rate. Adjusting weight decay in Adam provides regularization to prevent overfitting, and Adam can adaptively modify the learning rate. DK-STN is implemented using the PyTorch framework on the GTX 3090 24GB GPU.
%First, we split the global tropics into grid maps and cut the time dimension into time intervals (see Table~\ref{tab1}). Then, we select the reforecasting data, where each period contains 7 time intervals (i.e., 7 days). Further, we compute RMM indexes for each day as labels for supervised learning based on~\citep{b2}. In terms of $\mathcal{M}_S$, the convolution kernel sizes are set to $7 \times 7$ and $3 \times 3$. In terms of $\mathcal{M}_T$, the length of LSTM and hidden feature size in Encoder are set to 7 and 256 while in Decoder are set to 35 and 256 (i.e., $k=7$, $n=35$). \textcolor{red}{We choose $n=35$ days to facilitate comprehensive comparison, as ECMWF demonstrates high skill for 34 days in winter (details in Fig.~\ref{fig8} and Fig.~\ref{fig9}). Furthermore, we find that reducing to $n=30$ brings minor performance gain. So we use $n=35$ to facilitate comparison between models.} As for model training, the epoch size, \textcolor{red}{weight decay, and learning rate are set to 25, 0.001, and $1 \times e^{-4}$, respectively. We set Adam as our optimizer and it includes parameters, weight decay and learning rate. Adjusting the weight decay in Adam enables regularization to prevent overfitting. Furthermore, Adam can adaptively change the learning rate.} DK-STN is implemented with the Pytorch framework on the GTX 3090 24GB GPU.  

\noindent\textbf{Baseline Models.} We compare DK-STN with: (i) NWP methods. We choose CMA and ECMWF as they are common NWP methods, and ECMWF achieves state-of-the-art performance for MJO forecasts. (ii) Simple ANN methods. FNN~\citep{b19} and AR-RNN~\citep{b9} are simple attempts at deep learning on MJO forecasts. (iii) Spatio-temporal neural network methods. We chose CNN+LSTM and ConvLSTM as they are remarkable spatio-temporal neural network methods. Here, we only use our datasets to train and infer CNN+LSTM and ConvLSTM, while for other methods, we only select their best results for comparison. And prediction skill is acquired by adopting evaluation metrics (including COR and RMSE).

\begin{table*}[!t]
\begin{center}
\vspace{-0.2cm}
\caption{Datasets}
\begin{tabular}{c c c c c }
\hline
\textbf{Data sources} & \textbf{Variables} & \textbf{Spatial range} & \textbf{Temporal range} & \textbf{Train data \& Valid data} \\
\hline
ERA5 & \begin{tabular}[c]{@{}l@{}}OLR, SST \\ U200, U850\end{tabular} & \begin{tabular}[c]{@{}l@{}}$2.5^{\circ} \times 2.5^{\circ}$ resolution.\\ $15^{\circ}$N - $15^{\circ}$S latitude.\\ $0^{\circ}$ - $360^{\circ}$ longitude.\end{tabular} & \begin{tabular}[c]{@{}l@{}}Jan. 1, 1950 - Dec. 31, 2021 \\ Daily (averaged over 24 hours)\end{tabular} & \begin{tabular}[c]{@{}l@{}}Jan. 1, 1950 - Dec. 31, 2019 \\ Jan. 1, 2020 - Dec. 31, 2020\end{tabular} \\
%\hline
CMA$^{\left[1\right]}$& \begin{tabular}[c]{@{}l@{}}OLR, SST \\ U200, U850\end{tabular} & \begin{tabular}[c]{@{}l@{}}$2.5^{\circ} \times 2.5^{\circ}$ resolution.\\ $15^{\circ}$N - $15^{\circ}$S latitude.\\ $0^{\circ}$ - $360^{\circ}$ longitude.\end{tabular}& \begin{tabular}[c]{@{}l@{}}Jan. 1, 2006 - Dec. 31, 2021 \\ Daily\end{tabular} & \begin{tabular}[c]{@{}l@{}}Jan. 1, 2006 - Dec. 31, 2019 \\ Jan. 1, 2020 - Dec. 31, 2020\end{tabular} \\
%\hline
ECMWF$^{\left[1\right]}$ & \begin{tabular}[c]{@{}l@{}}OLR, SST \\ U200, U850\end{tabular} & \begin{tabular}[c]{@{}l@{}}$2.5^{\circ} \times 2.5^{\circ}$ resolution.\\ $15^{\circ}$N - $15^{\circ}$S latitude.\\ $0^{\circ}$ - $360^{\circ}$ longitude.\end{tabular}& \begin{tabular}[c]{@{}l@{}}Jan. 1, 2001 - Dec. 31, 2021 \\ Daily\end{tabular} & \begin{tabular}[c]{@{}l@{}}Jan. 1, 2001 - Dec. 31, 2019 \\ Jan. 1, 2020 - Dec. 31, 2020\end{tabular} \\ 
\hline
\multicolumn{5}{l}{\begin{tabular}[c]{@{}l@{}}This is the set of datasets that we utilized, and the model was tested with ERA5 data from Jan. 1, 2021, to Dec. 31, 2021. \\$~^{\left[1\right]}$ is from the S2S database.\end{tabular}} 
\end{tabular}
\vspace{-0.5cm}
\label{tab1}
\end{center}
\end{table*}

\begin{figure*}[!t]
\centering
    \captionsetup[subfloat]{labelsep=none,format=plain,labelformat=empty}
            \begin{minipage}{0.245\linewidth}
                    \subfloat[\quad (a)]{\includegraphics[width =  \linewidth]{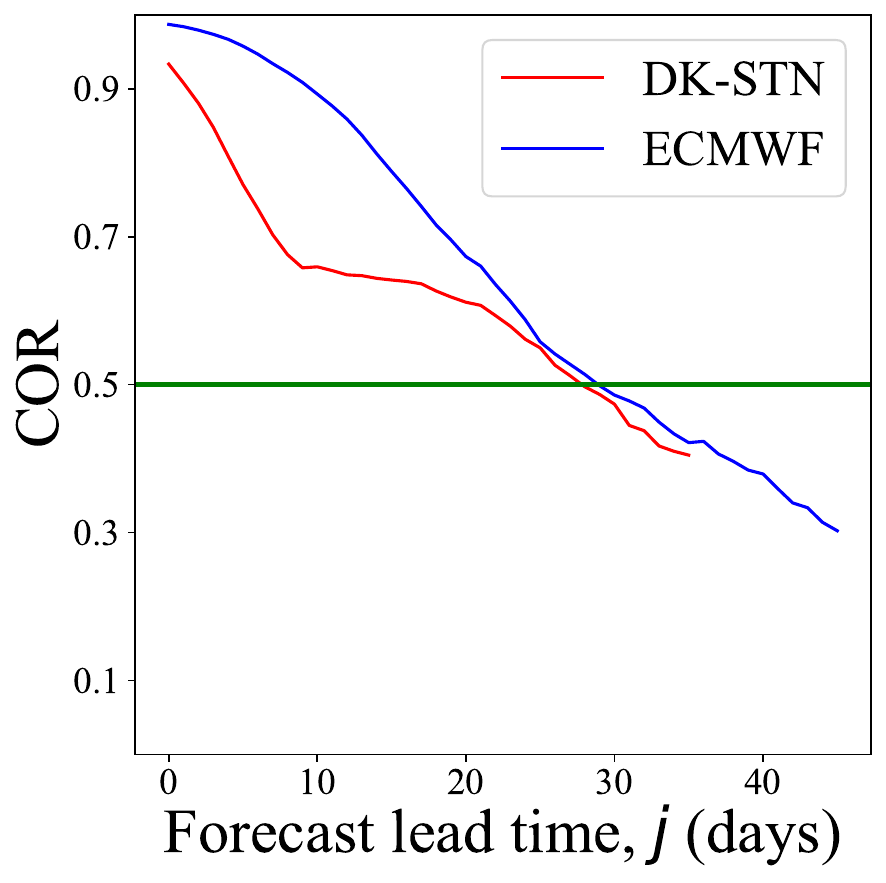}}\hfill
                    \label{6_a}
                    \subfloat[\quad (d)]{\includegraphics[width =  \linewidth]{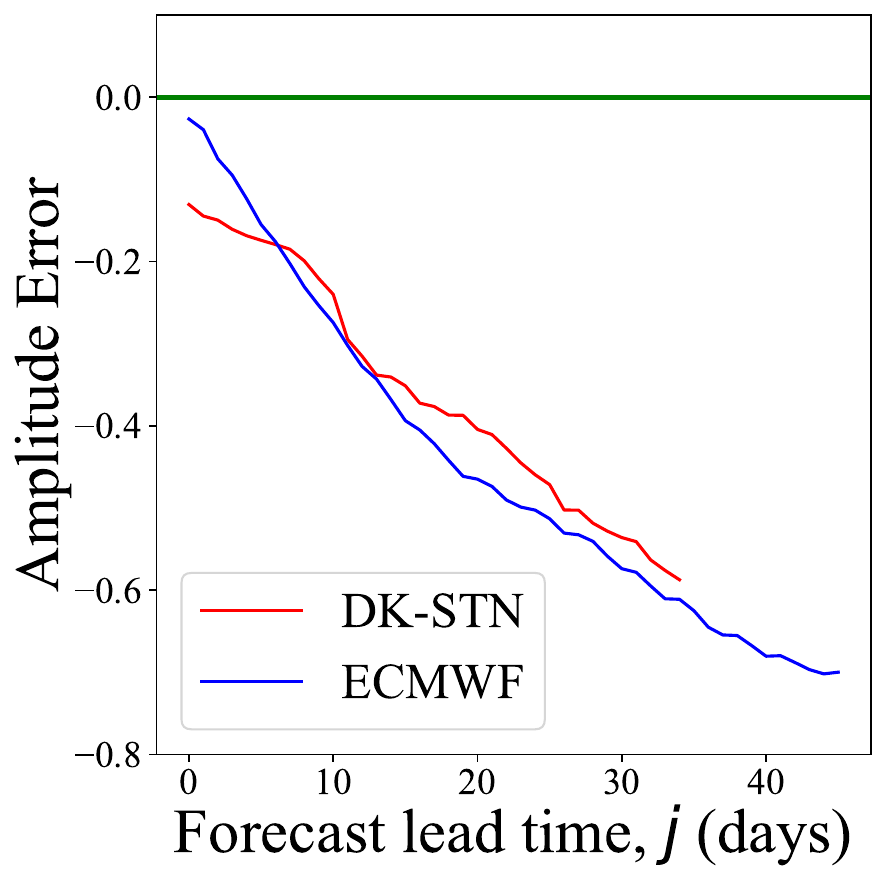}}\hfill
                    \label{6_d}
            \end{minipage}
            \begin{minipage}{0.245\linewidth}
                    \subfloat[\quad (b)]{\includegraphics[width =  \linewidth]{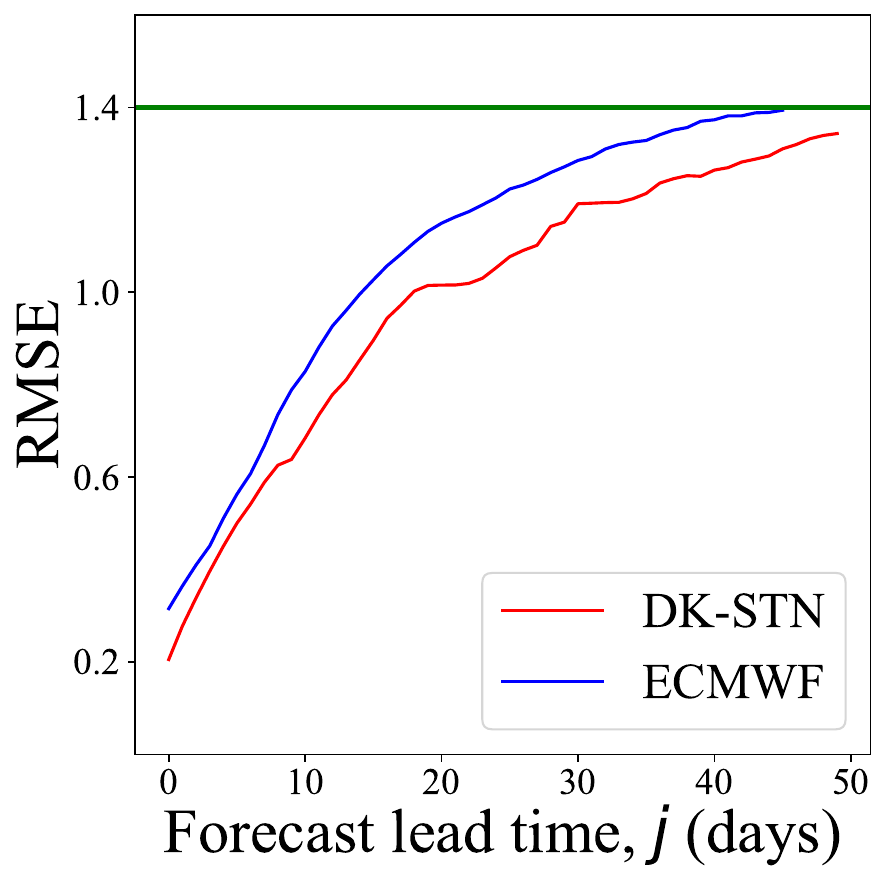}}\\
                    \label{6_b}
                    \subfloat[\quad (e)]{\includegraphics[width =  \linewidth]{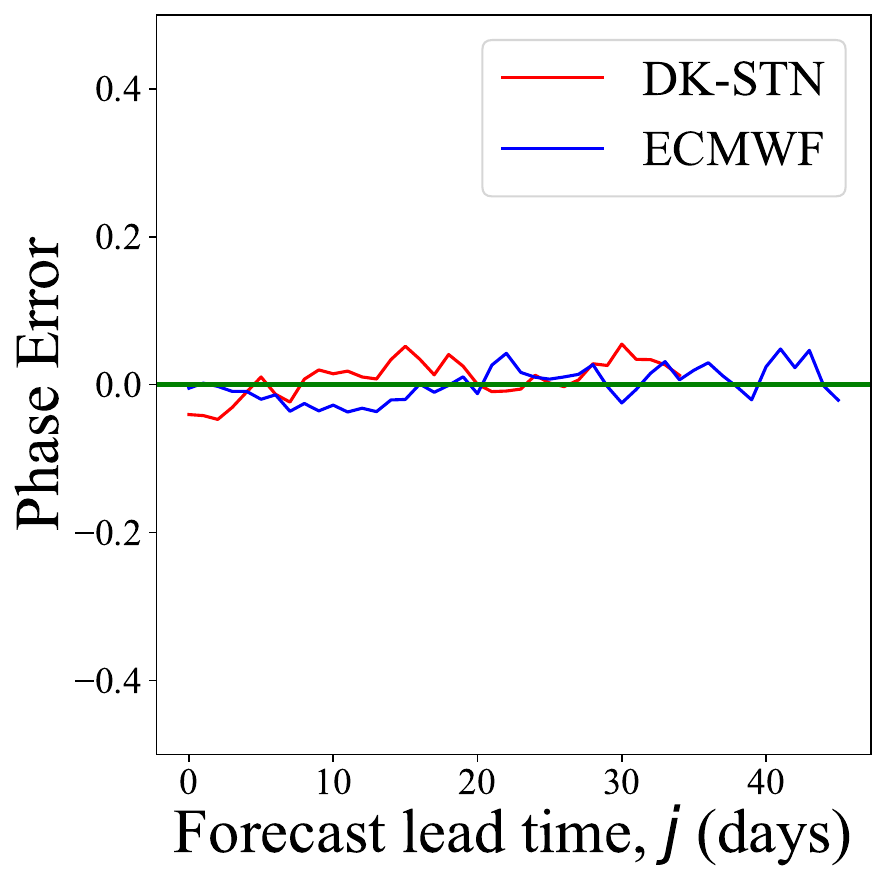}}
                    \label{6_e}
            \end{minipage}
            \begin{minipage}{0.49\linewidth}
                    \subfloat[\quad (c)]{\includegraphics[width =  \linewidth]{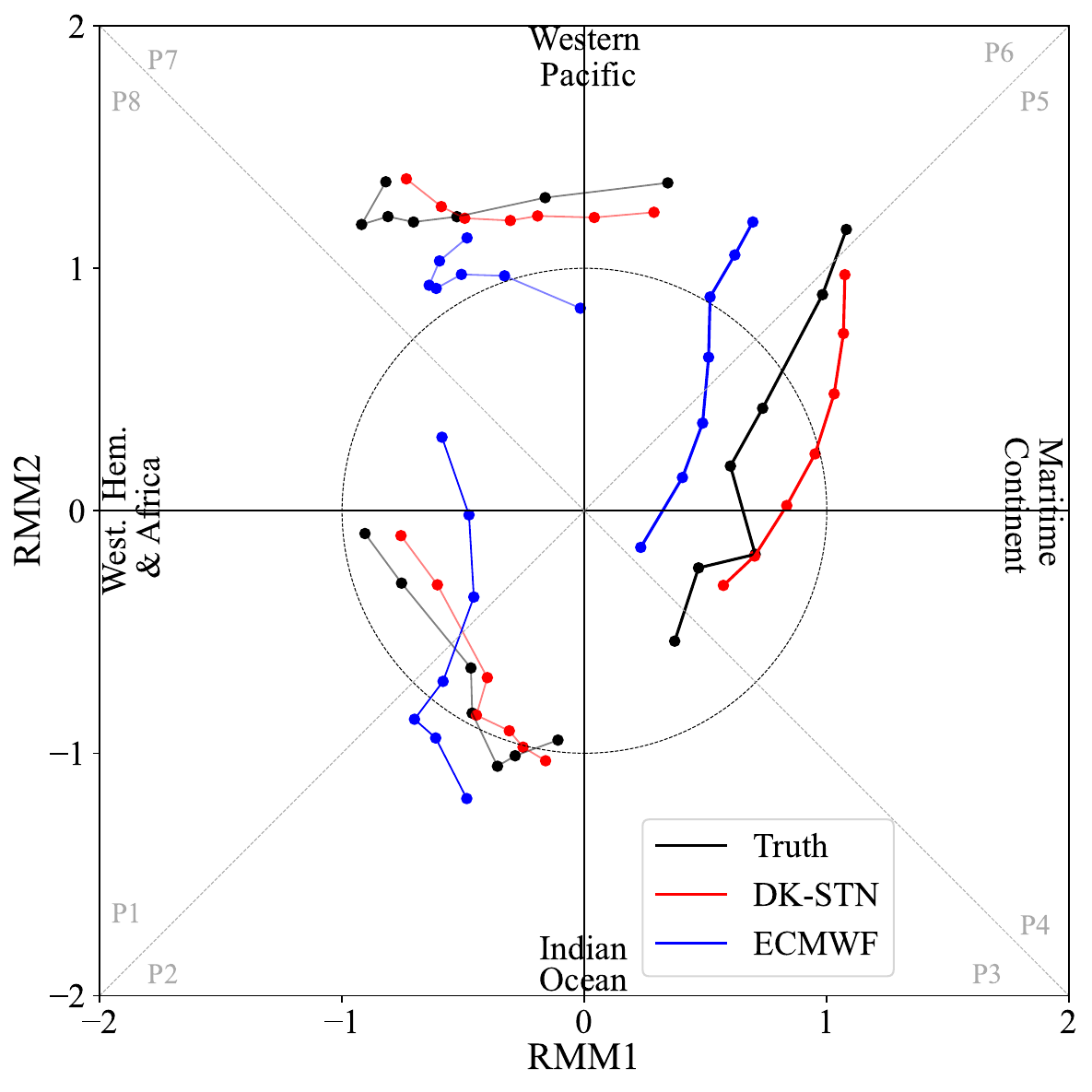}}
                    \label{6_c}
            \end{minipage}
		\caption{The comparison of DK-STN and ECMWF.}
  \vspace{-0.5cm}
		\label{fig6}
\end{figure*}

\begin{table*}[!t]
\vspace{-0.2cm}
\begin{center}
\caption{The Performance Comparison of All Methods}\label{tab2}
\begin{tabular}{c c c c c c } 
\hline
\multicolumn{2}{c}{All} & Input Data& \multicolumn{3}{c}{Prediction Skill (days)}\\ \cline{4-6}

\multicolumn{2}{c}{Methods} & (time length $k$) & ${\rm COR} \geq 0.50 $ & ${\rm RMSE} \leq 1.40$ & ${\rm COR} \geq 0.50 \& {\rm RMSE} \leq 1.40$ \\ 
\hline

NWP & CMA &- & 21 & 35 & 21\\ \cline{2-6} 
methods & ECMWF &- & 28 & 46 & 28\\ \cline{1-6}

Simple ANN & FNN &1$~^{\left[1\right]}$  & 17  & 20 & 17 \\ \cline{2-6} 

Methods & AR-RNN & 300$~^{\left[2\right]}$  & 25 & 40 & 25\\ \cline{1-6}

~ & CNN+LSTM &7$~^{\left[3\right]}$  & 15 & 21 & 15\\ \cline{2-6}

~ &ConvLSTM &7$~^{\left[3\right]}$  & 18 & 23 & 18\\ \cline{2-6}

Spatio-Temporal& STN &7$~^{\left[3\right]}$ & 18 & 35 & 18\\ \cline{2-6} 

Neural Network Methods & STN+DKEM &7$~^{\left[3\right]}$ & 23 & 45 & 23\\ \cline{2-6}

~ & STN+DKPM &7$~^{\left[3\right]}$ & 21 & 46 & 21\\ \cline{2-6}

~ & \textbf{DK-STN} &\textbf{7}$~^{\left[3\right]}$  & \textbf{28} & \textbf{51} & \textbf{28}\\ 
\hline
\multicolumn{6}{l}{\begin{tabular}[c]{@{}l@{}}$~^{\left[1\right]}$ Input circulation variable data for OLR, U200, U850, and SST for 1 day. (i.e., ${\rm input} \in \mathcal{R}^{l \times w \times 4}$) \\$~^{\left[2\right]}$ Input $RMM_1$ and $RMM_2$ for the first 300 days. (i.e., ${\rm input} \in \mathcal{R}^{300 \times 2}$) \\ $~^{\left[3\right]}$ Input circulation variable data for OLR, U200, U850, and SST for the first 7 days. (i.e., ${\rm input} \in \mathcal{R}^{7 \times l \times w \times 4}$)\end{tabular}} 
\end{tabular}
\vspace{-0.5cm}
\label{tab2}
\end{center}
\end{table*}

\subsection{Accuracy Study}\label{5.2}

\noindent\textbf{Overall Comparison.} As shown in Table~\ref{tab2}, we compare the optimal prediction skills of various methods. NWP methods consistently demonstrate robust performance, with ECMWF, in particular, achieving reliable MJO forecasts for up to 28 days and is the state-of-the-art. In contrast, simple ANN methods exhibit either low accuracy or significant limitations. FNN~\citep{b19} is restricted to a 17-day prediction horizon, while AR-RNN~\citep{b9} extends to 25 days but requires a 300-day input, reflecting  inefficiency.
%As shown in Table~\ref{tab2}, we compare the best prediction skills of different methods. NWP methods continue to produce strong results, particularly ECMWF, which can reliably forecast the MJO for up to 28 days and is the state-of-the-art, while simple ANN methods are either lowly accurate or have substantial limits. FNN~\citep{b19} can only predict 17 days, while AR-RNN~\citep{b9} can predict 25 days but needs 300 days of input. 
%NWP methods still have good results, while deep learning methods are either less effective or have significant limitations. FNN can only predict 17 days, while AR-RNN can predict 25 days but needs 300 days of input. And deep spatio-temporal neural networks have made outstanding progress as the network can capture spatio-temporal information of data to a certain extent. In contrast, the performance of DK-STN is comparable to ECMWF, reaching the state-of-the-art.

In terms of spatio-temporal neural network methods, they have made outstanding progress as they have been able to capture some spatio-temporal information about MJO. We select two of the most typical spatio-temporal neural networks, i.e., CNN+LSTM and the ConvLSTM mentioned in Section~\ref{DSTNN}, to perform MJO forecasting after inputting the same datasets as our model. It can be shown that the basic spatio-temporal neural network model STN we designed beats both networks, with strength prediction RMSE exceeding both and location prediction COR exceeding CNN+LSTM. While our model STN matches ConvLSTM in COR, it exhibits higher efficiency (refer to Section~\ref{5.5}). Moreover, with the incorporation of domain knowledge, our models significantly surpass both benchmarks, and the performance of DK-STN rivals ECMWF at the state-of-the-art level.
%And our models significantly outperform both after embedding domain knowledge, where the performance of DK-STN is even comparable to ECMWF at the state-of-the-art level.

\noindent\textbf{Compare COR and RMSE with ECMWF.} We further chose the same dataset to compare DK-STN and ECMWF. As depicted in Fig.~\ref{fig6}, we can find that under the same dataset of corresponding dates, the location predictions (i.e., COR) of DK-STN and ECMWF for MJO are roughly the same in the (a) figure. Because of the training with ECMWF data, DK-STN's predictions are similar to ECMWF's forecasting pattern which shows a rapid decline in the early days, followed by a slower decline. In the initial 10 days, DK-STN learns rapidly, showcasing its adeptness in capturing intricate patterns. Moreover, DK-STN demonstrates adaptive learning, identifying issues at the inflection point, resulting in a shift in convergence direction and slow down beyond 10 days.

And the strength predictions (i.e., RMSE) of DK-STN are better than ECMWF in the (b) figure. The prediction range of RMSE in Fig.~\ref{fig6} (b) (the same situation is in Fig.~\ref{fig7}, Fig.~\ref{fig8}, and Fig.~\ref{fig9}) beyond 50 days because we added new LSTMS blocks by copying the parameters of the last $n^{th}$ LSTMS in the trained DK-STN model. This allows iterative prediction outputs for increased duration to enable full comparison with ECMWF (up to 47 days). The first 35 days' predictions remain the same due to the iterative method.

\noindent\textbf{Compare $\overline{\rm AE}$ and $\overline{\rm PE}$ with ECMWF.} Furthermore, we evaluated the amplitude and phase errors of MJO predictions by DK-STN and ECMWF compared to the ground truth values, as illustrated in Fig.~\ref{fig6} (d) and (e). In Fig.~\ref{fig6} (d) both models exhibit negative $\overline{\rm AE}$, indicating that their predicted amplitudes are smaller than the true amplitudes. These errors tend to increase over longer prediction horizons. Although DK-STN has higher errors in the first 7 days, its $\overline{\rm AE}$ becomes lower than ECMWF thereafter. For medium and long-term MJO predictions, the performance of our DK-STN model is slightly superior to ECMWF. This implies that DK-STN excels in predicting moderate or strong MJO events compared to ECMWF.

Regarding phase errors in Fig.~\ref{fig6} (e), $\overline{\rm PE}$ fluctuates between positive and negative values, suggesting that the MJO propagation speeds predicted by both models are sometimes faster and sometimes slower than the truth. Numerically, their averaged 35-day $\overline{\rm PE}$ values are close, with -0.00823145 for ECMWF and 0.0083855 for DK-STN. This indicates comparable performance between DK-STN and ECMWF in MJO phase prediction.

\noindent\textbf{MJO Event Forecasting Analysis. } Since detecting and predicting MJO events poses challenges, we perform an additional MJO event forecasting study to validate the effectiveness of our proposed DK-STN by comparing with ECMWF. Specifically, we applied 7-day circulation data from Aug 7-14 2020, Nov 7-14 2020, Jan 8-15 2021 respectively to hindcast future RMM indexes using DK-STN for 35 days. As plotting all 35 days would be messy, we show the first 7 days to present the success of predicted MJO occurrences. Two MJO events happened during this period as shown in the lower two phase maps of Fig.~\ref{fig6} (c).

It can be observed that DK-STN successfully predicted the MJO occurrences. Additionally, compared to the prediction results of ECMWF, the DK-STN results are closer to the truth, i.e., the RMSE is lower, which is consistent with Fig.~\ref{fig6} (b). This analysis shows that DK-STN can effectively extract useful information for improved MJO predictions through domain knowledge infusion and spatio-temporal learning.
%In addition, we visualize several typical MJO events in Fig.~\ref{fig6} (c) using the spatial phase map of RMM indexes. When compared to the prediction results of ECMWF, the DK-STN results are closer to the observations, i.e., the RMSE is lower, \textcolor{red}{which is consistent with Fig.~\ref{fig6} (b)}.

\subsection{Ablation Study}\label{5.3}

\begin{figure}[!t]
\vspace{-0.5cm}
\captionsetup[subfloat]{labelsep=none,format=plain,labelformat=empty}
\centering
\subfloat[\quad (a)]{\includegraphics[width = .50 \linewidth]{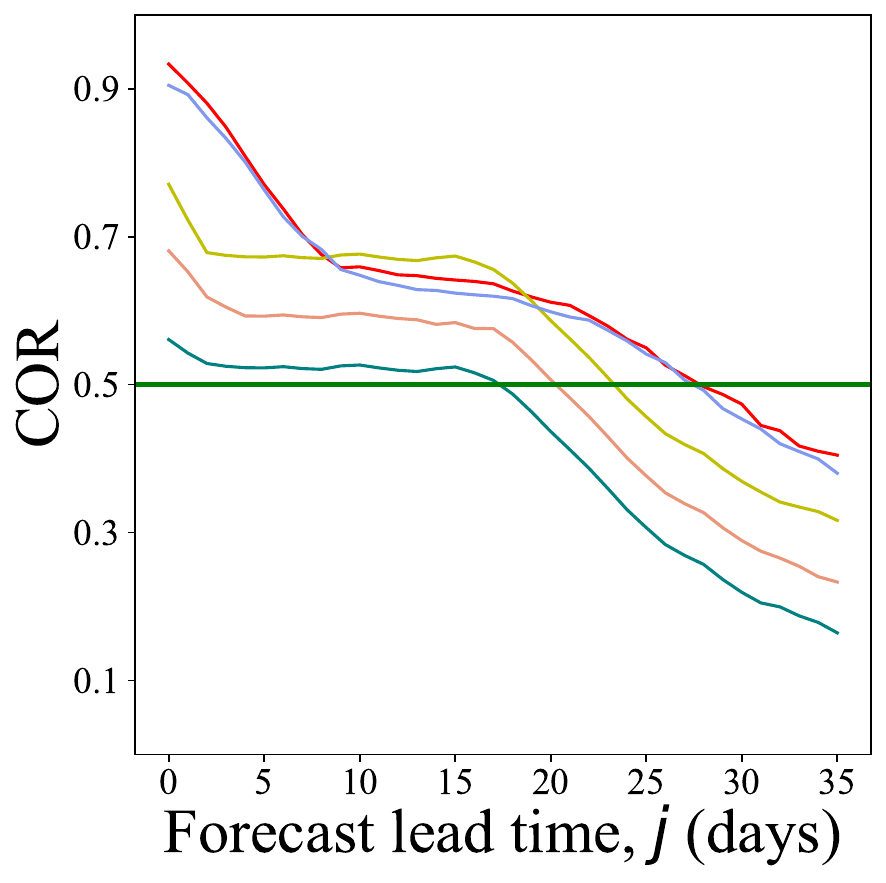}}\hfill
\subfloat[\quad (b)]{\includegraphics[width = .50 \linewidth]{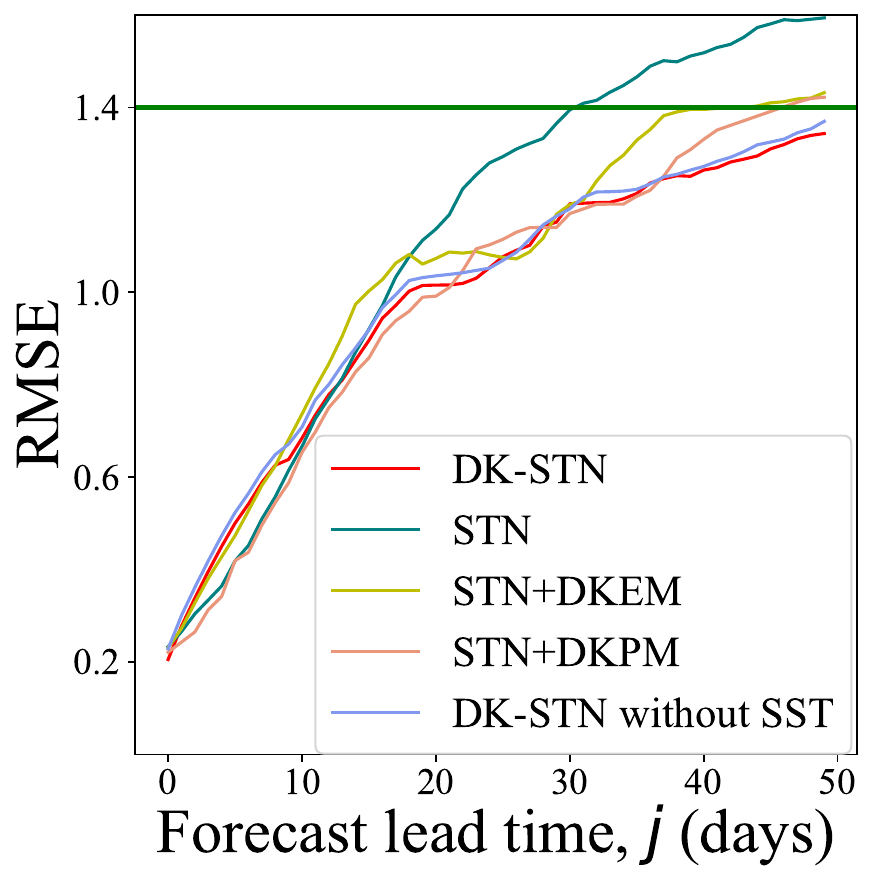}}
\caption{Ablation Study.}
\vspace{-0.5cm}
\label{fig7}
\end{figure}

For assessing the efficacy of the domain knowledge enhancement method (DKEM) and the domain knowledge processing method (DKPM), we conducted ablation experiments. These involved integrating each method individually into the foundational spatio-temporal network model, STN, used for MJO forecasting. The purpose was to evaluate their impact on forecasting accuracy. The results, depicted in Fig.~\ref{fig7} and summarized in Table\ref{tab2}, show that both domain knowledge methods contribute to enhanced MJO forecasting accuracy. The incorporation of DKEM led to a 5-day increase in prediction skill, while the inclusion of DKPM resulted in a 3-day improvement, underscoring the effectiveness of these domain knowledge methods.
%For evaluation of the domain knowledge enhancement method and the domain knowledge processing method, i.e., DKEM and DKPM, we implemented ablation experiments by integrating the two methods into the basic spatio-temporal network model STN for MJO forecasting, respectively, in order to test their efficiency. There is significant evidence that both domain knowledge methods can increase the accuracy of MJO forecasting to various degrees, as illustrated in Fig.~\ref{fig7} and Table~\ref{tab2}. The prediction skill increased by 5 days after adding DKEM and by 3 days after adding DKPM, demonstrating the effectiveness of domain knowledge methods.

Moreover, upon integrating both DKEM and DKPM into spatio-temporal neural network, i.e., DK-STN, we observed that the enhancement in prediction skills was not additive. In other words, it did not result in an increase of 8 days but rather contributed to a change of 10 days. This indicates a synergistic effect between the two methods, where their combined application led to a more substantial improvement in prediction skills than the sum of their individual impacts.
%Furthermore, we find that when both methods were implemented into neural networks, i.e., DK-STN, the improvement in prediction skills was not cumulative, meaning it was not an increase of 8 days but led to a change of 10 days. This demonstrated that both methods had a mutually reinforcing effect on each other.

Additionally, the addition of SST as an input variable to DK-STN is an important point of domain knowledge integration. To analyze its effect, we design comparative experiments between the DK-STN model and DK-STN without SST. As shown in Fig.~\ref{fig7}, their performance is similar. Specifically, compared to DK-STN without SST, DK-STN has slightly higher COR (increased by 0.02) and lower RMSE (reduced by 0.01). This shows the limited influence of 7-day SST data on prediction accuracy.

However, we find that DK-STN without SST takes more iterations (average 40 epochs) to converge at the comparable performance as the DK-STN model (average 25 epochs). This is because SST contributes to confining the variance of the overall data distribution, limiting likelihood values within a certain range. As the solution space gets narrowed down faster, the model is able to converge quicker towards likelihood extremum. In this way, the involvement of SST enables faster convergence during model training.

\subsection{Stability Study from Seasons}\label{5.4}

\begin{figure*}[!h]
  \centering
\captionsetup[subfloat]{labelsep=none,format=plain,labelformat=empty}
        \subfloat[\quad ECMWF (a)]{\includegraphics[width = .245 \linewidth]{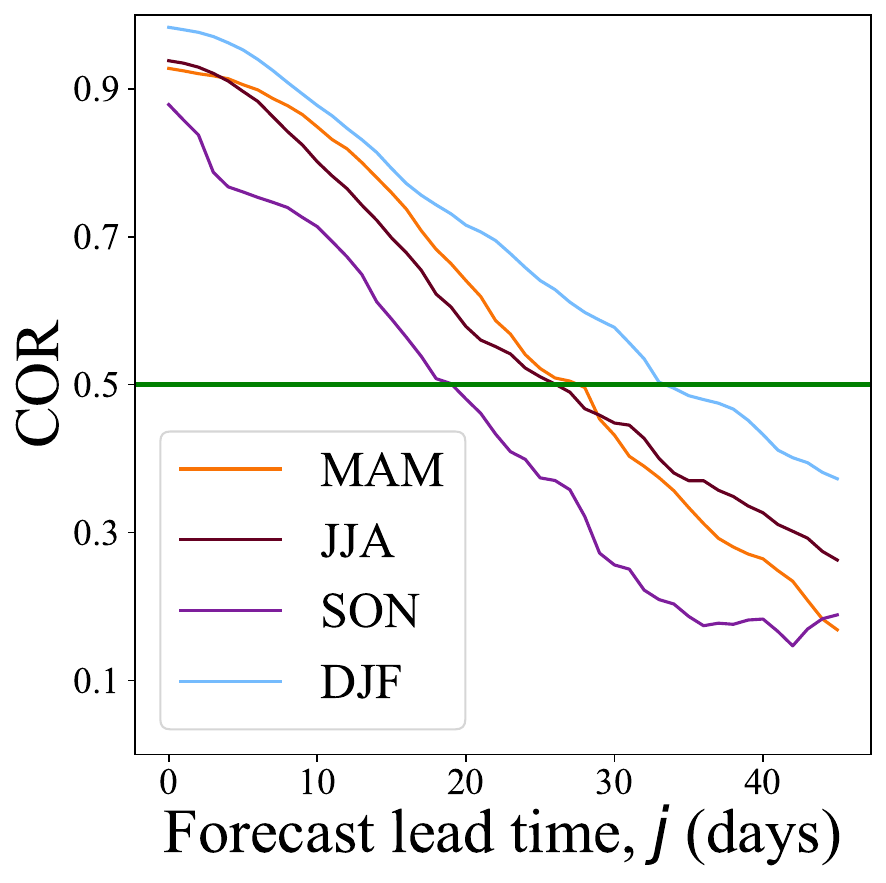}}\hfill
        \subfloat[\quad ECMWF (b)]{\includegraphics[width = .245 \linewidth]{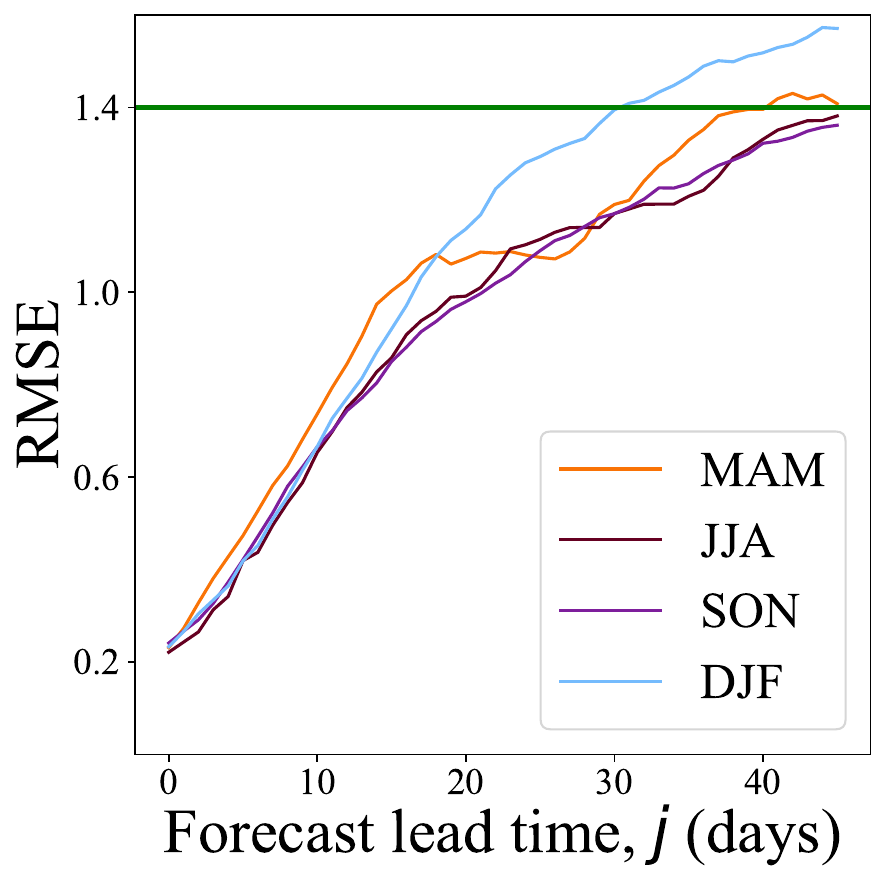}}\hfill
        \subfloat[\quad DK-STN (a)]{\includegraphics[width = .25 \linewidth]{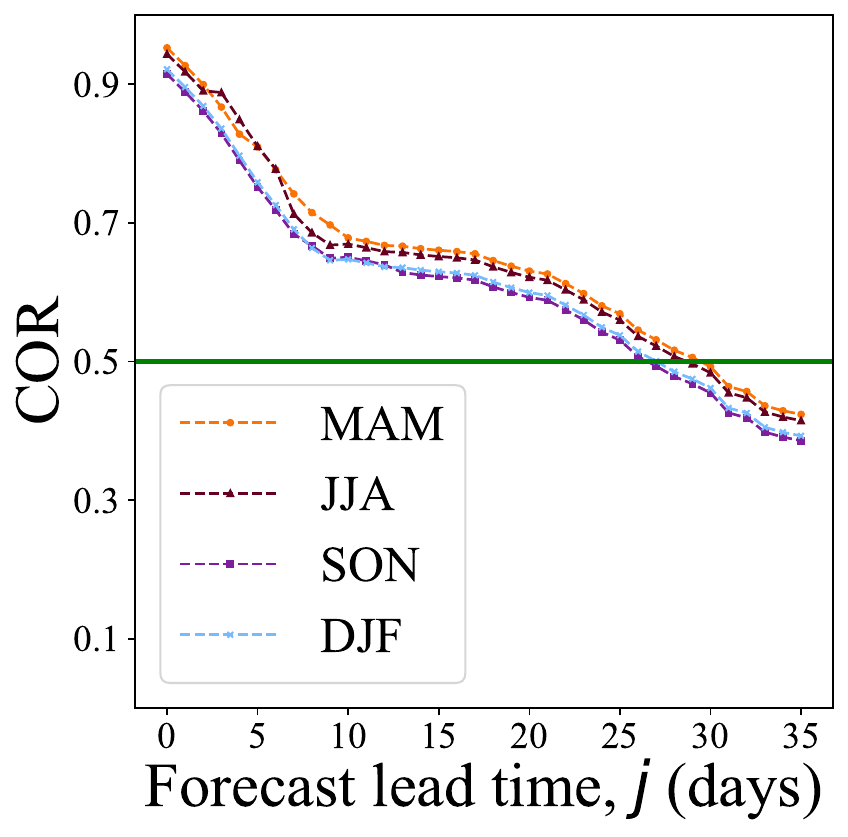}}\hfill
        \subfloat[\quad DK-STN (b)]{\includegraphics[width = .25 \linewidth]{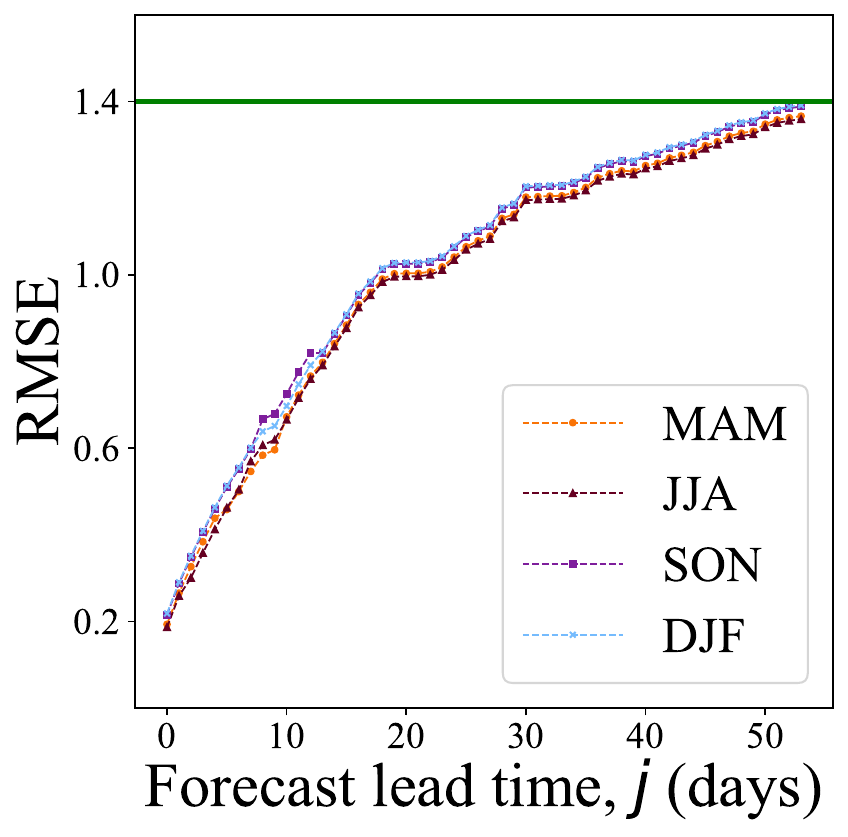}}
		\caption{Prediction skills during different seasons.}
  \vspace{-0.5cm}
  \label{fig8}
\end{figure*}

\begin{figure*}[!t]
\centering
\captionsetup[subfloat]{labelsep=none,format=plain,labelformat=empty}
\subfloat[\quad MAM (a)]{\includegraphics[width = .25 \linewidth]{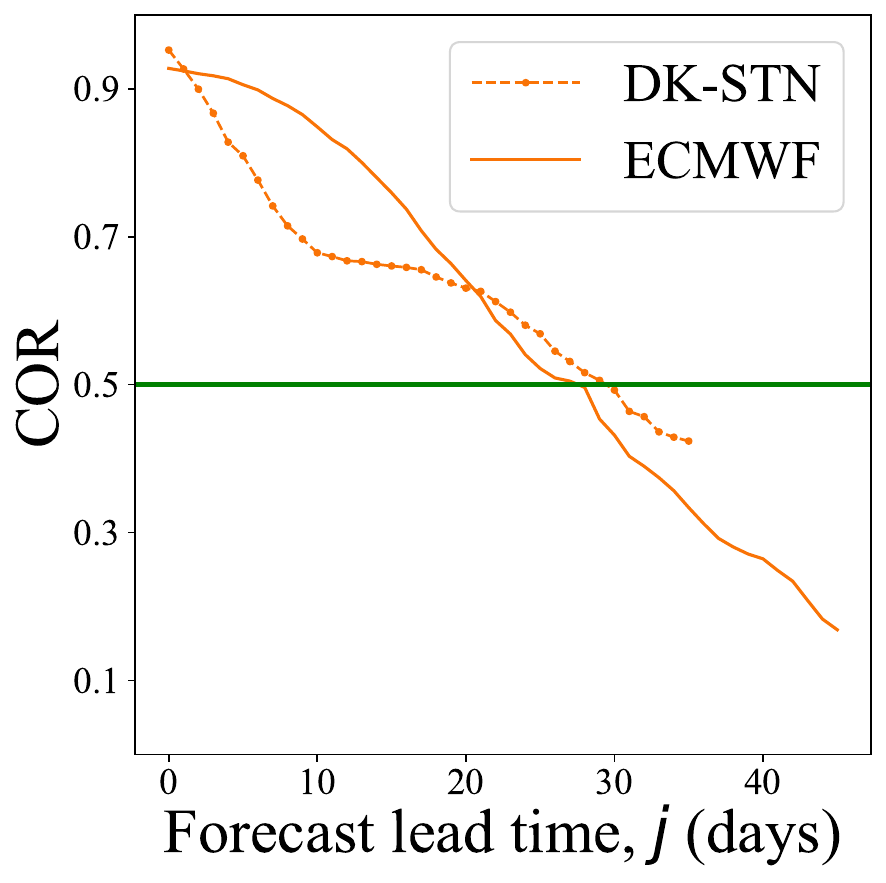}}\hfill
\subfloat[\quad JJA (a)]{\includegraphics[width = .25 \linewidth]{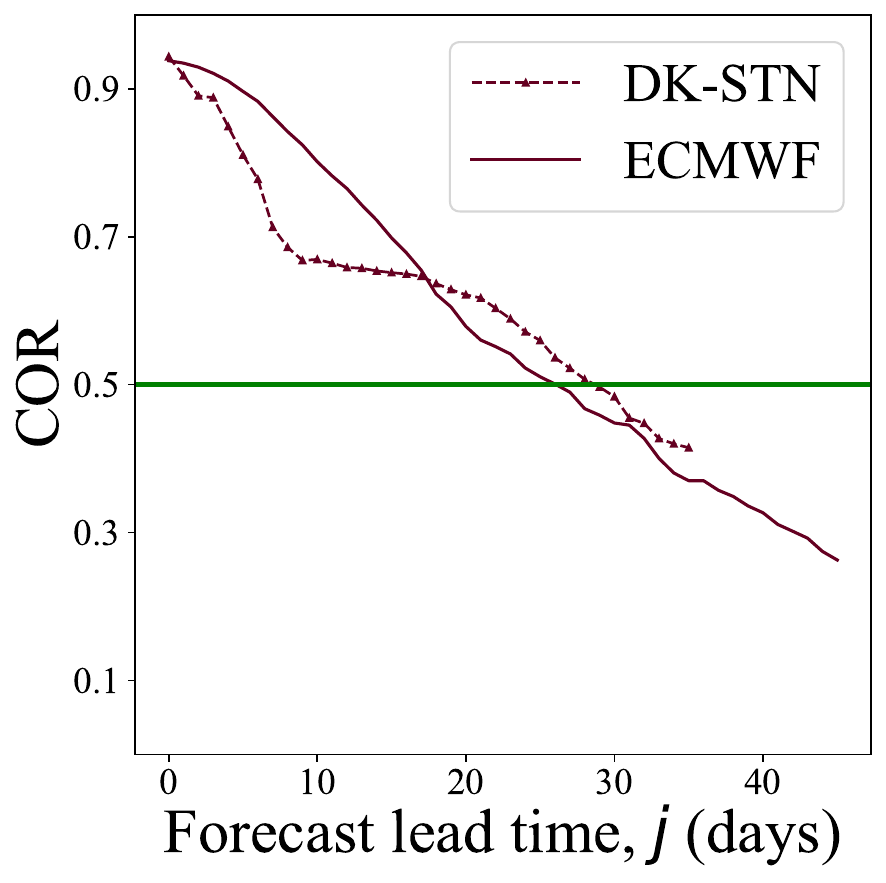}}\hfill
\subfloat[\quad SON (a)]{\includegraphics[width = .25 \linewidth]{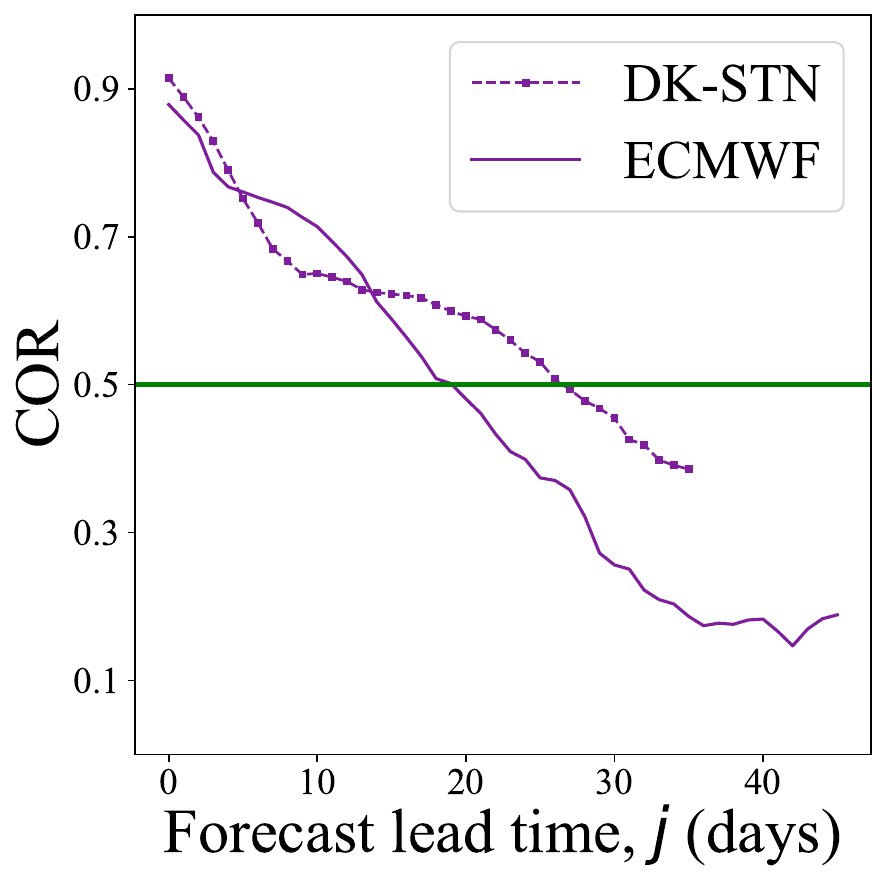}}\hfill
\subfloat[\quad DJF (a)]{\includegraphics[width = .25 \linewidth]{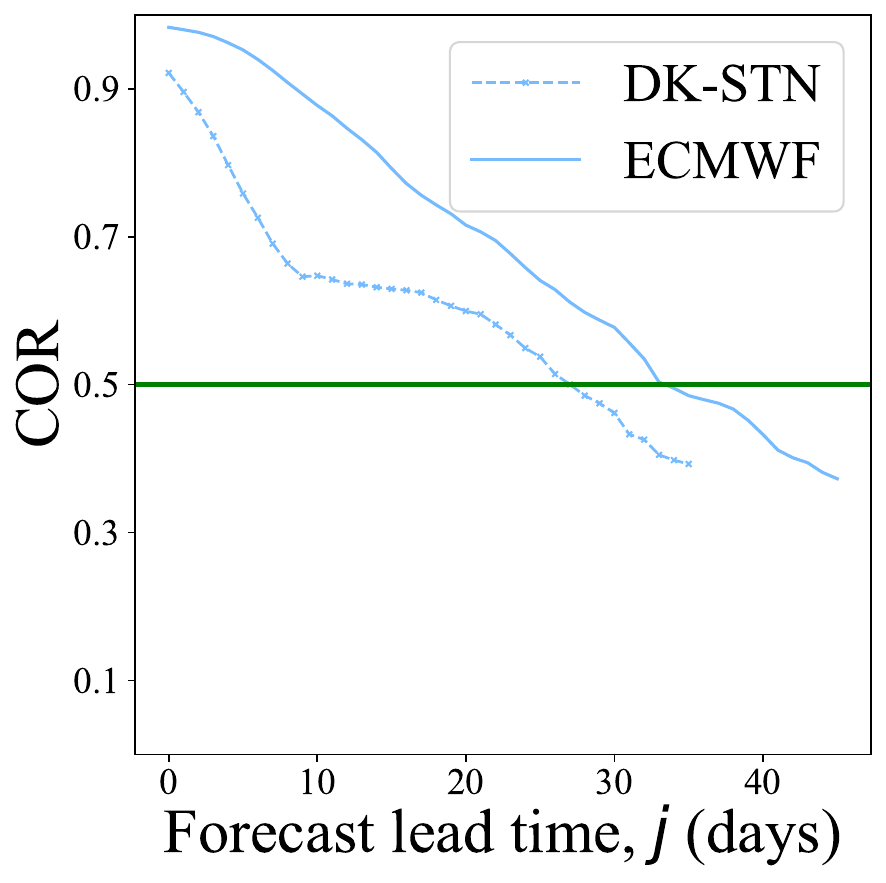}}\\ \vspace{-4mm}
\subfloat[\quad MAM (b)]{\includegraphics[width = .25 \linewidth]{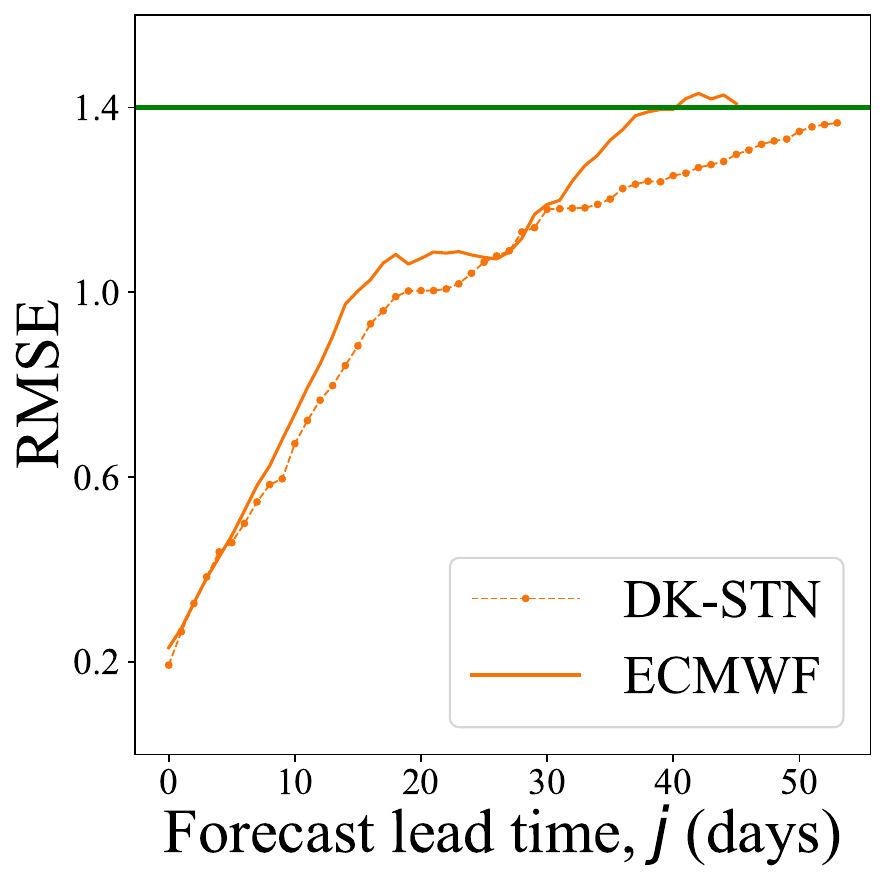}}\hfill
\subfloat[\quad JJA (b)]{\includegraphics[width = .25 \linewidth]{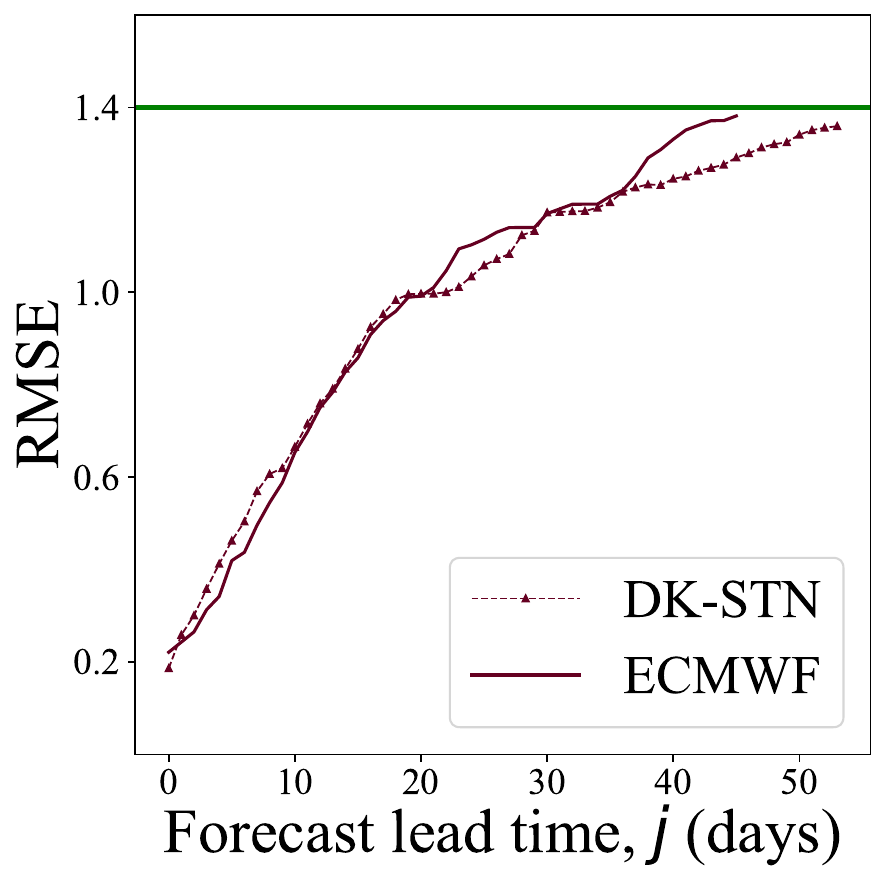}}\hfill
\subfloat[\quad SON (b)]{\includegraphics[width = .25 \linewidth]{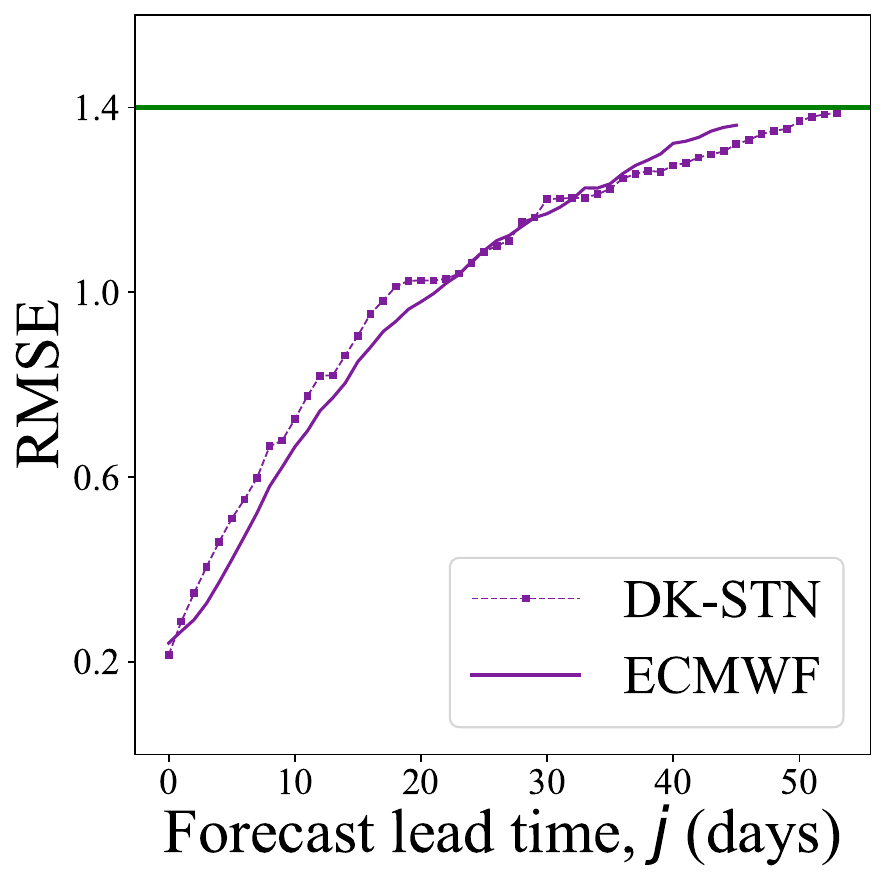}}\hfill
\subfloat[\quad DJF (b)]{\includegraphics[width = .25 \linewidth]{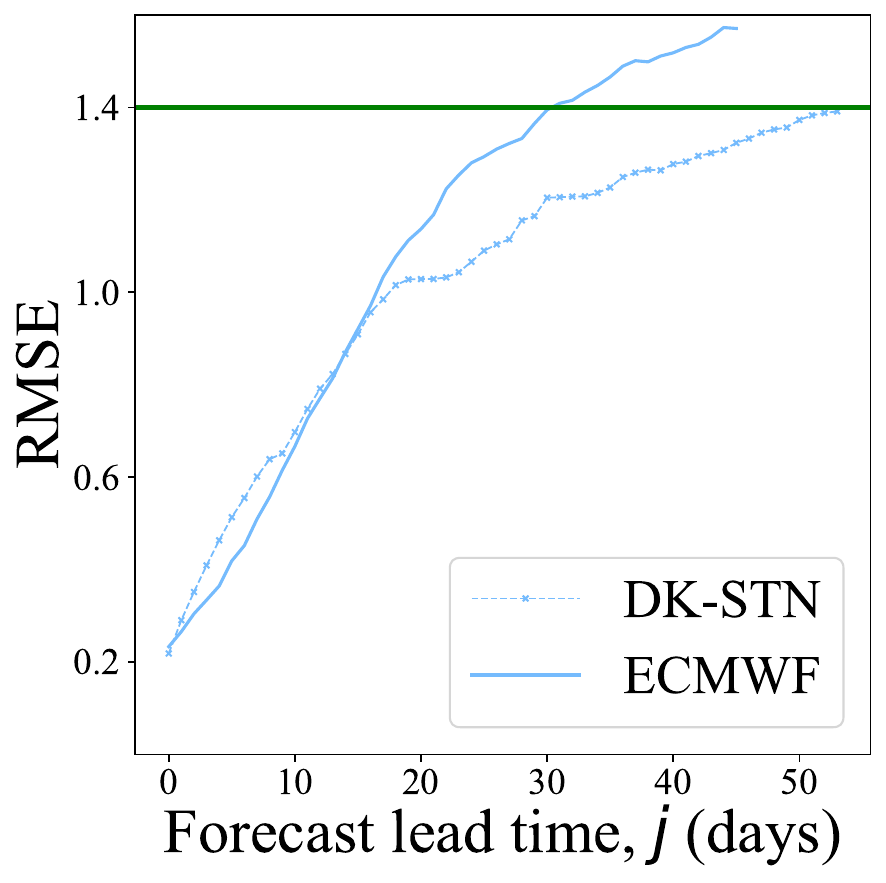}}
\caption{The comparison of prediction skills between DK-STN and ECMWF in different seasons.}
\vspace{-0.5cm}
\label{fig9}
\end{figure*}
In accordance with relevant meteorological studies, MJO forecasts using NWP methods are sensitive to seasonal variations, resulting in less stable predictions across different seasons~\citep{unstable,b19}. As a general rule, MJO events are best predicted during the winter months. 
Researchers believe this trend is attributed to the heightened activity of MJO during northern hemisphere winters.  To validate our model’s stability from seasons, we apply DK-STN to analyze each season within the dataset, benchmarking its performance against the ECMWF in different seasons. The results are shown in Fig.~\ref{fig8} and Fig.~\ref{fig9}.
%In accordance with relevant meteorological studies, MJO's prediction abilities are not as stable as those associated with the change of seasons. As a general rule, MJO events are best predicted during the winter months. Researchers believe this is likely due to the fact that the MJO is most active during the winter months in the northern hemisphere. We use DK-STN to perform the same analysis for each season in the dataset and compare it to the state-of-the-art method ECMWF. Fig.~\ref{fig8} and Fig.~\ref{fig9} depict the results.

Fig.~\ref{fig8} respectively depicts the prediction skills of ECMWF and DK-STN methods in different seasons: MAM (i.e., March-April-May) is spring, JJA (i.e., June-July-August) is summer, SON is autumn, and DJF is winter. As observed, the prediction skills of MJO by ECMWF are very unstable in different seasons, which is proved by previous studies~\citep{b6,b7,b9,b19}. In terms of prediction skill, the most accurate time is DJF (i.e., December-January-February), and it is roughly 34 days. The most difficult is SON (i.e., September-October-November), which is only 23 days. However, the design of the DK-STN exhibits superior stability in different seasons. Its prediction skills in different seasons are roughly the same, with only 2–3 days of error.

As shown in Fig.~\ref{fig9}, we perform a comprehensive comparison of prediction skills between ECMWF and DK-STN across various seasons. In terms of RMSE, DK-STN consistently outperforms ECMWF throughout all seasons. Regarding COR, DK-STN surpasses ECMWF in spring, summer, and autumn, with particularly noteworthy improvement in autumn. Although DK-STN exhibits a slight lag behind ECMWF in winter, this difference doesn't compromise its overall competitive edge. The overall COR of DK-STN is comparable to that of ECMWF. This demonstrates that our model ensures both stability and cutting-edge accuracy in prediction performance.
%As shown in Fig.~\ref{fig9}, we further compare the prediction skills of ECMWF and DK-STN in each season. For RMSE, DK-STN significantly outperforms ECMWF in each season, and for COR, DK-STN is better than ECMWF in spring, summer, and autumn, where it is obviously better in autumn and only worse than ECMWF in winter, which not only ensures the stability of prediction performance but also the state-of-the-art accuracy.

\subsection{Efficiency Study}\label{5.5}

\begin{figure}[!t]
\vspace{-0.5cm}
\captionsetup[subfloat]{labelsep=none,format=plain,labelformat=empty}
\centering
\subfloat[\quad (a)]{\includegraphics[width = .49 \linewidth]{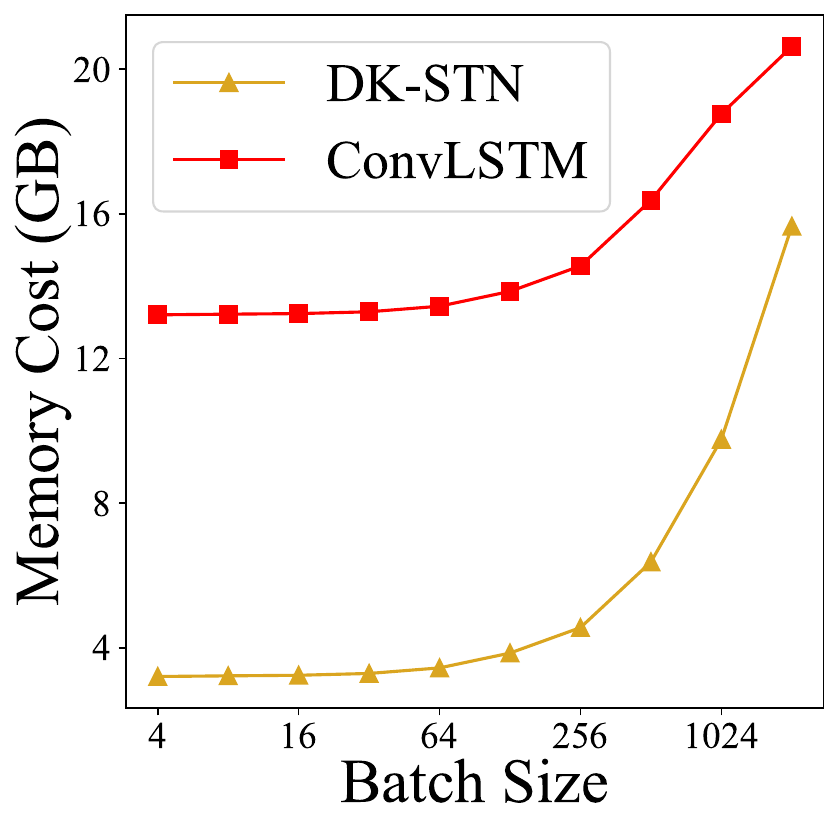}}\hfill
\subfloat[\quad (b)]{\includegraphics[width = .50 \linewidth]{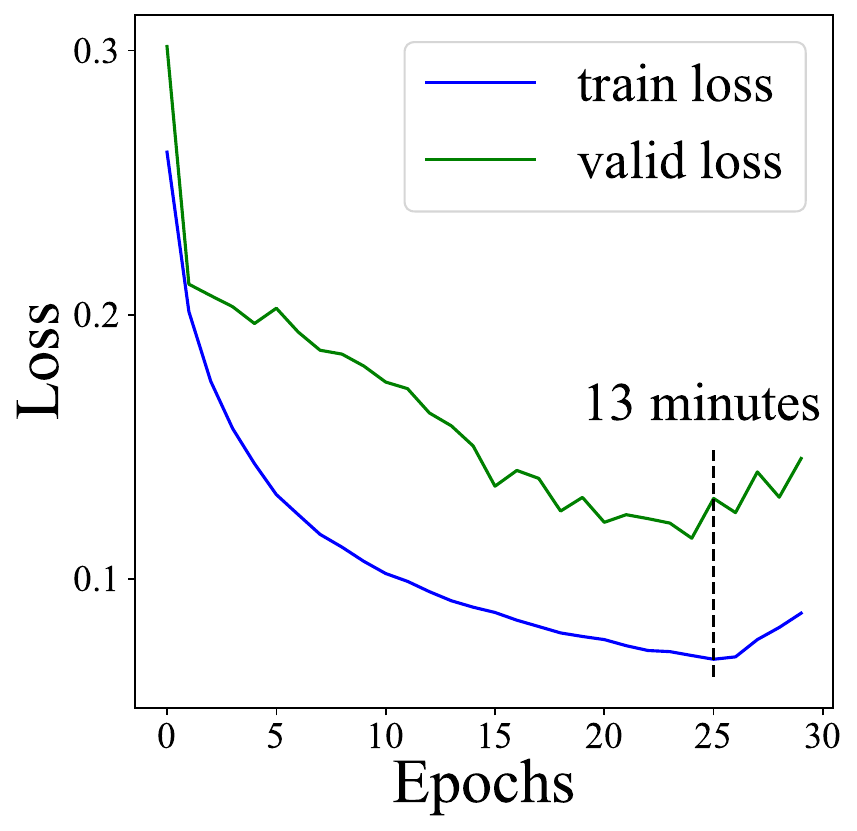}}
\caption{Efficiency Study.}
\vspace{-0.5cm}
\label{fig10}
\end{figure}

To evaluate the efficiency of our model, we focus on two aspects. Firstly, in terms of training memory, as illustrated in Fig.~\ref{fig10} (a), we compare DK-STN with ConvLSTM using the same training datasets. We exclude CNN+LSTM from this comparison due to its notably low accuracy, despite its simple network structure that doesn't demand substantial memory. As shown in Fig.~\ref{fig10} (a), we can find DK-STN surpasses ConvLSTM significantly in terms of memory occupation, regardless of the batch size.
%Last but not least, we study the model's efficiency. In terms of training memory, as depicted in Fig.~\ref{fig10} (a), we compare DK-STN to ConvLSTM with the same training datasets. We do not compare CNN+LSTM in this case because its accuracy is too low. This is despite the fact that its network structure is too simple for it to acquire a great deal of memory. In terms of memory occupied, our spatio-temporal neural network model DK-STN outperforms ConvLSTM by a wide margin, regardless of the size of the batch.

In terms of training speed, the loss variation of the neural network when the model achieves the fastest convergence (i.e., batch size = 16) is shown in Fig.~\ref{fig10} (b). We can see it converge quickly. In particular, within a period of 25 epochs (corresponding to 13 minutes), the training and valid losses first decrease rapidly and then subside to a minimum over time. This running-time efficiency is acceptable in real-life deployments. 

In addition, our model also clearly outperforms the NWP methods on memory occupy and training speed as NWP methods spend a lot to solve complex differential equations by a high-performance computing cluster. In contrast, our model is a lightweight and fast MJO forecasting model, designed for efficient and rapid predictions.

\section{Conclusion and Discussion}\label{sec6}
In this paper, we propose a model based on circulation variables for MJO forecasting, called the Domain Knowledge Embedded Spatio-temporal Network (DK-STN). This model ensures forecast stability and efficiency while improving prediction accuracy by embedding domain knowledge. In the data part of the model, a domain knowledge enhancement method is used to provide a variety of data to support network training. Additionally, in order to ensure stable and efficient MJO forecasts, our approach combines a spatial residual convolution module with a temporal attentive adaptation module, i.e., STN. Meanwhile, we add a domain knowledge processing module to increase the capture rate of MJO information. And this module is aimed at improving the accuracy of MJO as well. Our total forecasting results not only outperform both existing ANN methods and the state-of-the-art NWP method (operational forecasts from ECMWF) demonstrated in our experiments but also provide high stability and efficiency.

%\textcolor{red}{We explore the impact of different prediction horizons 'n' on model forecasting. We set $n$ to 30 days and retrained the DK-STN model. The refined model ($n=30$) outperforms its predecessor with $n=35$, achieving superior 29-day accurate MJO predictions on the test set. We believe the shortened $n$ reduces the probability of averaging distribution bias, focusing on the prediction target, and mitigating bias exposure. Simultaneously, this adjustment expedites gradient descent convergence towards the likelihood extremum. Our exploration showcases the model's capability for accurate predictions across flexible time scales.}

In future work, we will investigate other factors that influence the MJO in order to increase the model's prediction accuracy even further. Meanwhile, we will also attempt to apply the model to other typical climate events.

\printcredits

\section*{Acknowledgments}

This work was supported by the Natural Science Foundation of Jilin Province (Grant 20230101062JC), the National Key Research and Development Plan of China (Grant 2017YFC1502306), and the National Natural Science Foundation of China (No. 42175052, No. 61902143, No. 62272190, and No. U19A2061).

\bibliography{ref}

\end{document}